\documentclass[runningheads]{llncs}

\usepackage{eccv}

\PassOptionsToPackage{table}{xcolor}
\usepackage{eccvabbrv}

\usepackage{graphicx}
\usepackage{booktabs}
\usepackage{rotating}

\usepackage[accsupp]{axessibility}

\usepackage[square,sort,comma,numbers]{natbib}
\usepackage{pdflscape}
\usepackage{multirow}
\usepackage{colortbl}
\usepackage{color}
\usepackage{pgfplots}
\usepackage{amsmath}
\usepackage{tikz}
\usepackage{sidecap}
\usepackage{floatrow}
\usepackage{caption}
\usepackage{subcaption}
\usetikzlibrary{arrows,shapes,backgrounds,patterns,fadings,decorations.pathreplacing,decorations.pathmorphing}
\usepackage{orcidlink}
\usepackage{hyperref}

\newcommand{\vect}[1]{\boldsymbol{#1}}
\def\mog{MDN}
\newcommand{\ours}{CARFF: Conditional Auto-encoded Radiance Field for 3D Scene Forecasting}
\newcommand{\oursshort}{CARFF}
\newfloatcommand{capbtabbox}{table}[][\FBwidth]
\definecolor{lightgreen}{rgb}{0.9, 1, 0.9}
\definecolor{regulargreen}{rgb}{0.0, 0.5, 0.0}
\newcolumntype{P}[1]{>{\centering\arraybackslash}p{#1}}

\begin{document}

\title{CARFF: Conditional Auto-encoded Radiance Field for 3D Scene Forecasting}

\author{Jiezhi ``Stephen'' Yang\inst{1\thanks{Core contributors}}\orcidlink{0000-0002-0135-2628} \and
Khushi Desai\inst{2\footnotemark[1]}\orcidlink{0009-0009-3024-8406} \and
Charles Packer\inst{3}\orcidlink{} \and Harshil Bhatia \inst{4}\orcidlink{0000-0002-4561-0860} \and Nicholas Rhinehart \inst{3}\orcidlink{0000-0003-4242-1236} \and Rowan McAllister \inst{5}\orcidlink{0000-0002-9519-2345} \and Joseph E. Gonzalez \inst{3}\orcidlink{0000-0003-2921-956X}}

\authorrunning{J. Yang et al.}

\institute{Harvard University MA 02138, USA \and
Columbia University NY 10025, USA \and
UC Berkeley CA 94720, USA \and Avataar.ai KA 560103, India \and Toyota Research Institute CA 94022, USA}

\maketitle

\begin{abstract}
    We propose \ours, a method for predicting future 3D scenes given past observations. Our method maps 2D ego-centric images to a distribution over plausible 3D latent scene configurations and predicts the evolution of hypothesized scenes through time. Our latents condition a global Neural Radiance Field (NeRF) to represent a 3D scene model, enabling explainable predictions and straightforward downstream planning. This approach models the world as a POMDP and considers complex scenarios of uncertainty in environmental states and dynamics. Specifically, we employ a two-stage training of Pose-Conditional-VAE and NeRF to learn 3D representations, and auto-regressively predict latent scene representations utilizing a mixture density network. We demonstrate the utility of our method in scenarios using the CARLA driving simulator, where \oursshort{} enables efficient trajectory and contingency planning in complex multi-agent autonomous driving scenarios involving occlusions. Video and code are available at \href{https://stephenyangjz.github.io/carff_website}{www.carff.website}.
\end{abstract}

\section{Introduction}
\label{sec:intro}
Humans often imagine what they cannot see given partial visual context.
Consider a scenario where reasoning about the unobserved is critical to safe decision-making: for example, a driver navigating a blind intersection.
An expert driver will plan according to what they believe may or may not exist in occluded regions of their vision.
The driver's belief -- defined as the understanding of the world modeled with consideration for inherent environment uncertainties -- is informed by their partial observations (i.e., the presence of other vehicles on the road), as well as their prior knowledge (e.g., past experience navigating this intersection).

When reasoning about the unobserved, humans form complex beliefs about the existence, position, shapes, colors, and textures of occluded scene portions (e.g., an oncoming car). 
Autonomous systems with high-dimensional sensor data, like video or LiDAR, traditionally reduce this data to low-dimensional state information (e.g., position and velocity of tracked objects) for prediction and planning.

In addition to tracking fully observed objects, this object-centric framework handles partially observed settings by considering potentially dangerous unobserved objects. These systems often plan for worst-case scenarios, such as a "ghost car" at the edge of the visible field of view \cite{Tas_2018}.
\begin{figure}
\includegraphics[width=0.6 \textwidth]{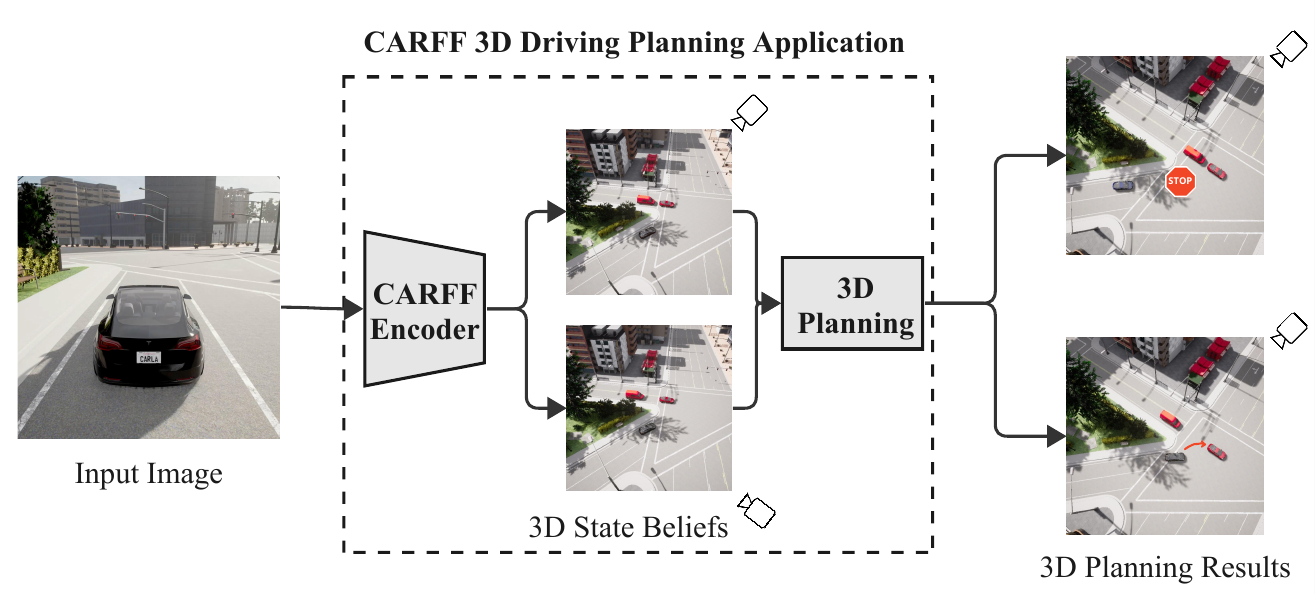}
\caption{\textbf{CARFF 3D planning application for driving}. An input image containing a partially observable view of an intersection is processed by CARFF's encoder to establish 3D environment state beliefs, i.e. the predicted possible state of the world: whether or not there could be another vehicle approaching the intersection. These beliefs are used to forecast the future in 3D for planning, generating one among two possible actions for the vehicle to merge into the other lane.}
\label{fig:teaser}
\end{figure}

Recent advances in neural rendering, particularly Neural Radiance Fields (NeRF), have significantly improved 3D scene representation learning. 
NeRF enables novel view synthesis, thus simplifying the process of viewing behind occlusions. NeRF decouples the dependancy of scene representation from traditional object detection and tracking, allowing for the capture of vital visual information that might be missed by detectors, yet is crucial for safe decision-making.
NeRF's implicit density representation of explicit geometry also facilitates its direct application in motion planning without the need for rendering.
NeRF's ability to represent both visual and geometric information makes them a more general and intuitive 3D representation for autonomous systems.

Despite NeRF's advantages, achieving probabilistic predictions in 3D based on reasoning from occluded views is challenging.
For example, discriminative models that yield categorical predictions are unable to capture the underlying 3D structure, impeding their ability to model uncertainty.
While prior work on 3D representation captures view-invariant structures, their application is primarily confined to simple scenarios \cite{kosiorek2021nerfvae}. 
We present \oursshort{}, which to our knowledge, is the first forecasting approach in scenarios with partial observations that uniquely facilitates stochastic predictions in a partially observable Markov decision process (POMDP) within a 3D representation, effectively integrating visual perception and geometry.
Specifically, we make the following contributions:
\begin{enumerate}
    \item We propose a novel architecture \emph{PC-VAE}: Pose-Conditioned Variational Autoencoder. The encoder maps potentially partially observable ego-centric images to pose-invariant latent scene representations, which hold state beliefs of the POMDP with implicit probability distributions (see Sec.~\ref{sec:nerf}).
    
    \item We develop the two-stage training pipeline that uniquely enables complex scene modeling with a probabilistic objective. This involves separately training the PC-VAE and a latent conditioned neural radiance field that functions as a 3D decoder, enabling interpretable predictions (see Sec.~\ref{sec:nerf}).
    
    \item We design a mixture density model to predict the evolution of 3D scenes over time stochastically and regressively in the encoder belief space (see Sec.~\ref{sec:mdn}). This allows for an effective sampling based-controller to output actions in the POMDP.
\end{enumerate}

We demonstrate how \oursshort{} can be used to enable contingency planning in complex driving scenarios that require reasoning into visual occlusions on CARLA simulated datasets inspired by autonomous driving planning tasks~\cite{9564424, packer2022is, 10229239, Yu_2019}.  A potential application of \oursshort{} is illustrated in Fig.~\ref{fig:teaser}.
\section{Related work}
\label{related}
\subsection{NeRF and 3D representations}
\paragraph{Neural radiance fields.} Neural Radiance Fields (NeRF)~\cite{mildenhall2020nerf,tancik2020fourier,barron2021mip} for 3D representations generate high-resolution, photorealistic scenes. Instant Neural Graphics Primitive (Instant-NGP)~\cite{muller2022instant} speeds up training and rendering time by introducing a multi-resolution hash encoding.
Other works like Plenoxels~\cite{fridovich2022plenoxels} and DirectVoxGo (DVGO) ~\cite{sun2022direct} also provide similar speedups.
Recent advancements in volumetric representations such as 3D Gaussian Splatting~\cite{kerbl3Dgaussians} enhance rendering efficiency while maintaining compatibility with traditional NeRF applications~\cite{fei20243d}. We utilize Instant-NGP for its accessibility, although our approach is adaptable to alternative rendering methods.
NeRFs have also been extended for several tasks such as modeling large-scale unbounded scenes~\cite{barron2021mip, turki2022mega,tancik2022block}, scene from sparse views~\cite{truong2023sparf,deng2022depth,roessle2022dense} and multiple scenes~\cite{kosiorek2021nerfvae,wang2021ibrnet}.
For an in-depth survey on neural representation learning and its applications we refer the reader to~\cite{Tewari2022NeuRendSTAR}.
\begin{figure}
    % \vspace*{-\baselineskip}
    \centering
    \includegraphics[width= 0.4 \columnwidth]{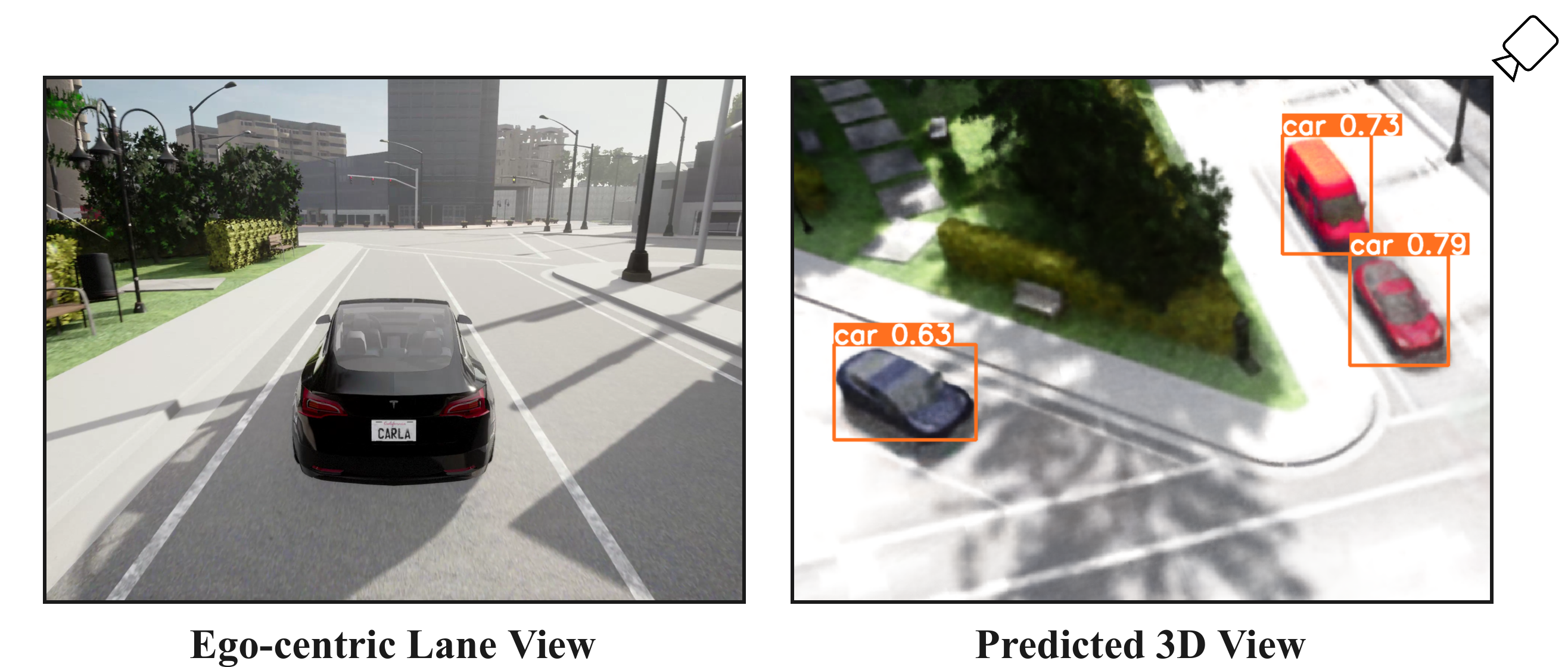}
    \caption{\textbf{Novel view planning application}. \oursshort{} allows reasoning behind occluded views from the ego car as simple as moving the camera to see the sampled belief predictions, allowing simple downstream planning using, for example, density probing or 2D segmentation models from arbitrary angles.}
    \label{fig:nvs}
\end{figure}

Generalizable novel view synthesis models, like pixelNeRF and pixelSplat~\cite{yu2021pixelnerf, charatan2023pixelsplat}, learn a scene prior to render novel views from sparse existing ones. In contrast, \oursshort{} is based on a VAE, encoding a probabilistic objective and decoding to future 3D scenes. Dynamic NeRF models scenes with moving or deforming objects, within which a widely used approach is to construct a canonical space and predict a deformation field~\cite{
liu2022devrf,park2021hypernerf,pumarola2020dnerf,park2021nerfies}.
The canonical space is usually a static scene, and the model learns an implicitly represented flow field~\citep{park2021nerfies,pumarola2020dnerf}. A recent line of work also models dynamic scenes via different representations and decomposition ~\cite{cao2023hexplane, shao2023tensor4d}.
These approaches tend to perform better for spatially bounded and predictable scenes with relatively small variations \cite{yu2021pixelnerf, luiten2023dynamic, park2021nerfies, cao2023hexplane}.
Moreover, these methods only solve for changes in the environment but are limited in incorporating stochasticity in the environment.
% \vspace*{-\baselineskip}
\paragraph{Multi-scene NeRF:}Our approach builds on multi-scene NeRF approaches~\cite{kosiorek2021nerfvae, xu2022point,wang2021ibrnet,tretschk2023scenerflow} that learn a global latent scene representation, which conditions the NeRF, allowing a single NeRF to effectively represent various scenes. A similar method, NeRF-VAE, was introduced by Kosiorek \textit{et al.}~\cite{kosiorek2021nerfvae} to create a geometrically consistent 3D generative model with generalization to out-of-distribution cameras. However, NeRF-VAE~\cite{kosiorek2021nerfvae} is prone to mode collapse when faced with complex visual information (see Sec.~\ref{sec:carff_eval}).
\subsection{Scene Forecasting}
\paragraph{Planning in 2D space:} 
Planning in large, continuous state-action spaces is challenging due to exponentially large search spaces~\cite{papadimitriou1987complexity}, leading to various approximation methods for tractability~\cite{pineau2003point,mcallister2017data}. Model-free~\cite{hausknecht2015deep,pan2017virtual,toromanoff2020end} and model-based~\cite{cao2021instance} reinforcement learning frameworks, along with other learning-based methods~\cite{codevilla2019exploring,packer2022is}, have emerged as viable approaches. Additionally, methods forecast for downstream control~\cite{ivanovic2021mats}, learn behavior models for contingency planning~\cite{rhinehart2021contingencies}, or predict the existence and intentions of unobserved agents~\cite{pmlr-v205-packer23a}. While these methods operate in 2D, we reason under partial observations and account for these factors in 3D.

\paragraph{NeRF in robotics:}
Recent works have applied NeRFs in robotics for localization~\cite{yen2021inerf}, navigation~\cite{adamkiewicz2022nerfplanning,Marza_2023_ICCV}, dynamics modeling~\cite{driess2022compnerfdyn,liu2022devrf},and robotic grasping~\cite{ichnowski2021dex,kerr2022evo}. Adamkiewicz \textit{et al.}\cite{adamkiewicz2022nerfplanning} propose quadcopter motion planning in NeRF models by sampling the learned density function, useful for forecasting and planning. Driess \textit{et al.}\cite{driess2022compnerfdyn} employ a graph neural network to learn dynamics in a multi-object NeRF scene. Li \textit{et al.}~\cite{li20223nerfdy} focus on pushing tasks and address grasping and planning with NeRF and a separate latent dynamics model.
Prior approaches work in simple, static scenes~\cite{adamkiewicz2022nerfplanning} or uses deterministic dynamics models~\cite{li20223nerfdy}. \oursshort{} addresses complex, realistic 
environments with both state and dynamics uncertainty, considering potential object existence and unknown movements.
\section{Method}
Recent advancements in 3D scene representation allow for modeling environments in a contextually rich and interactive 3D space. This offers analytical benefits, such as spatial analysis with soft occupancy grids and object detection through novel view synthesis.
Given these advantages, our primary objective is to develop a model for probabilistic 3D scene forecasting in dynamic environments.
However, direct integration of 3D scene representation via NeRF and probabilistic models like VAE often involves non-convex and inter-dependent optimization, which causes unstable training. 
For instance, NeRF's optimization may rely on the VAE's latent space being structured to provide informative gradients.
\begin{figure*}
    \centering
    \includegraphics[width= 0.9 \textwidth]{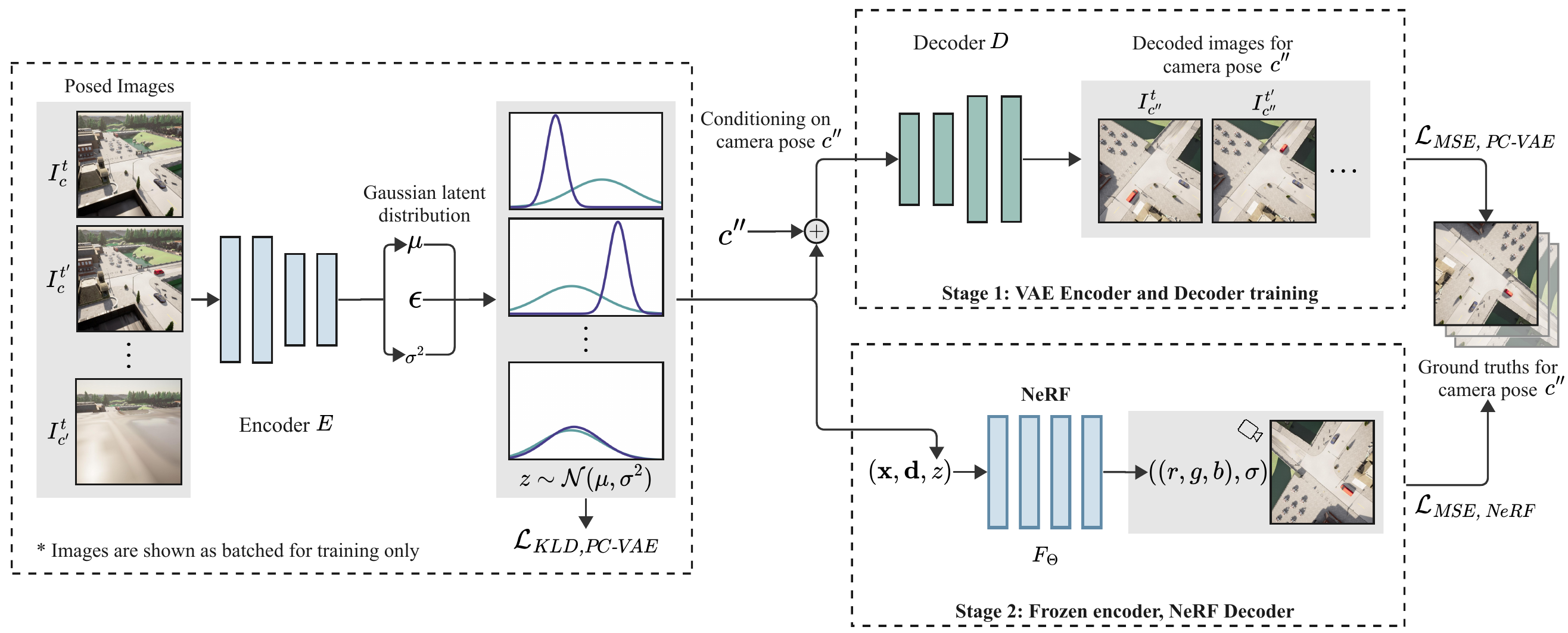}
    \caption{\textbf{Visualizing \oursshort{}'s two stage training process}. \textbf{Left:} The convolutional VIT-based encoder encodes each image $I$ at timestamps $t, t'$ and camera poses $c, c'$ into Gaussian latent distributions. Assuming two timestamps and an overparameterized latent, one Gaussian distribution will have a smaller $\sigma^2$, and different $\mu$ across timestamps. \textbf{Upper Right:} The pose-conditional decoder stochastically decodes the sampled latent $z$ using the camera pose $c''$ into images $I_{c''}^t$ and $I_{c''}^{t'}$. The decoded reconstruction and ground truth images are used for the loss $\mathcal{L_{\text{MSE, PC-VAE}}}$. \textbf{Lower Right:} A NeRF is trained by conditioning on the latent variables sampled from the optimized Gaussian parameters. These parameters characterize the distinct timestamp distributions derived from the PC-VAE. An MSE loss is calculated for NeRF as $\mathcal{L_{\text{MSE, NeRF}}}$.}
    \label{fig:vae}
\end{figure*}
To navigate these complexities, our method bifurcates the training process into two stages (see Fig.~\ref{fig:vae}). First, we train the PC-VAE to learn view-invariant scene representations. Next, we replace the decoder with a NeRF to learn a 3D scene from the latent representations. The latent scene representations capture the environmental states and dynamics over possible underlying scenes, while NeRF synthesizes novel views within the belief space, giving us the ability to see the unobserved (see Fig.~\ref{fig:nvs} and Sec.~\ref{sec:nerf}).
During prediction, uncertainties can be modeled by sampling latents auto-regressively from a predicted Gaussian mixture, allowing for effective decision-making. 
To this extent, we approach scene forecasting as a POMDP over latent distributions, which enables us to capture multi-modal beliefs for planning amidst perceptual uncertainty (see Sec.~\ref{sec:mdn}).
\subsection{Pose-Conditional VAE (PC-VAE) and NeRF}
\label{sec:nerf}
\paragraph{Architecture:}
\label{architecture}
We assume that the model follows a Markovian process, and thus each belief state only depends on the previous.
Given a scene $S_t$ at timestamp $t$, we have an ego-centric observation image $I^t_c$ captured from camera pose $c$.  
The objective is to formulate a 3D representation of the image that holds implicit probability distributions of the possible states, where we can perform a forecasting step that evolves the scene forward. Here, the POMDP can be seen as an MDP in belief space~\cite{KAELBLING199899}.
To achieve this, we utilize a radiance field conditioned on latent variable $z$ sampled from the posterior distribution $q_{\phi}(z | I_{c}^{t})$.
Now, to learn the posterior, we utilize PC-VAE. 
We construct an encoder using convolutional layers and a pre-trained ViT on ImageNet \cite{dosovitskiy2021image}.
The encoder learns a mapping from the image space to a Gaussian distributed latent space $ q_\phi (z|I^t_c) = \mathcal{N}(\mu,\sigma^2)$ parametrized by mean $\mu$ and variance $\sigma^2$. The decoder, $p(I|z,c)$, conditioned on camera pose $c$, maps the latent $z \sim \mathcal{N}(\mu,\sigma^2)$ into the image space $I$. This helps the encoder to generate latents that are invariant to the camera pose $c$.

To enable 3D scene modeling, we employ Instant-NGP~\cite{muller2022instant}, which incorporates a hash grid and an occupancy grid to enhance computation efficiency. Additionally, a smaller multilayer perceptron (MLP), $F_\theta(z)$ can be utilized to model the density and appearance, given by:
\begin{equation}
    F_\theta(z): (\mathbf{x},\mathbf{d}, z) \rightarrow ((r,g,b),\sigma)
\end{equation} 
Here, $\mathbf{x} \in \mathbb{R}^3$ and $\mathbf{d} \in (\theta, \phi)$ represent the location vector and the viewing direction respectively. The MLP is conditioned on the sampled scene latents $z\sim q_\phi(z | I_c^t)$ (see Appendix~\ref{sec:train_details}). 
\paragraph{Training methodology:}
\label{training_method}
The architecture alone does not enable us to model complex scenarios, as seen through a similar example in NeRF-VAE~\cite{kosiorek2021nerfvae}. 
A crucial contribution of our work is our two-stage training framework which stabilizes the training. First, we optimize the convolutional ViT based encoder and pose-conditional convolutional decoder in the pixel space for reconstruction.
This enables our method to deal with more complex and realistic scenes as the encoding is learned in a semantically rich 2D space. 
By conditioning the decoder on camera poses, we achieve disentanglement between camera view angles and scene context, making the representation view-invariant and the encoder 3D-aware.
Once rich latent representations are learned, we replace the decoder with a latent-conditioned NeRF over the latent space of the frozen encoder. 
The NeRF reconstructs encoder beliefs in 3D for novel view synthesis.
\paragraph{Loss:}
\label{VAE-loss}
PC-VAE is trained using standard VAE loss, with mean square error (MSE) and a Kullback–Leibler (KL) divergence given by evidence lower bound: 
    \begin{multline}
        \mathcal{L}_{\textit{PC-VAE}} = \mathcal{L}_{\textit{MSE, PC-VAE}} + \mathcal{L_{\textit{KLD, PC-VAE}}} = \\ || p(I|z,c'') -  I^t_{c''} \|^2  +  \mathbb{E}_{q(z|I^t_c)}[\log p(I|z)] - w_{\textit{KL}} D_{KL}(q_{\phi}(z|I^{t}_{c})\; || \; p(I | z))  
    \end{multline}
where $w_{\textit{KL}}$ denotes the KL divergence loss weight and $z~\sim~q_{\phi}(z|I^t_{c})$.
To make our representation 3D-aware, our posterior is encoded using camera $c$ while the decoder is conditioned on a randomly sampled pose $c''$.

KL divergence regularizes the latent space to balance conditioned reconstruction and stochasticity under occlusion. 
An elevated KL divergence loss weight $w_{\textit{KL}}$ pushes the latents closer to a standard normal distribution, $\mathcal{N}(0, 1)$, thereby ensuring probabilistic sampling in scenarios under partial observation. 
However, excessive regularization causes the latents to be less separable, leading to mode collapse. 
To mitigate this, we adopt delayed linear KL divergence loss weight scheduling to strike a balanced $w_{\textit{KL}}$.

Next, we learn a NeRF decoder on the posterior of the VAE to model scenes. At any timestamp $t$ we use a standard photometric loss for training the NeRF, given by the following equation:
\begin{equation}
   \mathcal{L_{\textit{MSE, NeRF}}} =  \| I^{t}_{c} - \textit{render}(F_\theta(\cdot| q_\phi(z | I^t_{c}))) \|^2
\end{equation}
We use a standard rendering algorithm as proposed by M{\"u}ller \textit{et al.}~\cite{muller2022instant}. 
Next, we build a forecasting module over the learned latent space of our pose-conditional encoder.
\subsection{Scene Forecasting}
\label{sec:mdn}
\paragraph{Formulation:} 
The current formulation allows us to model scenes with different configurations across timestamps. 
In order to forecast future configurations of a scene given an ego-centric view, we need to predict future latent distributions.
We formulate the forecasting as a POMDP over the posterior distribution $q_\phi(z | I^t_{c})$ in the PC-VAE's latent space.

\begin{figure*}
    \centering
    \includegraphics[width= 0.95\textwidth]{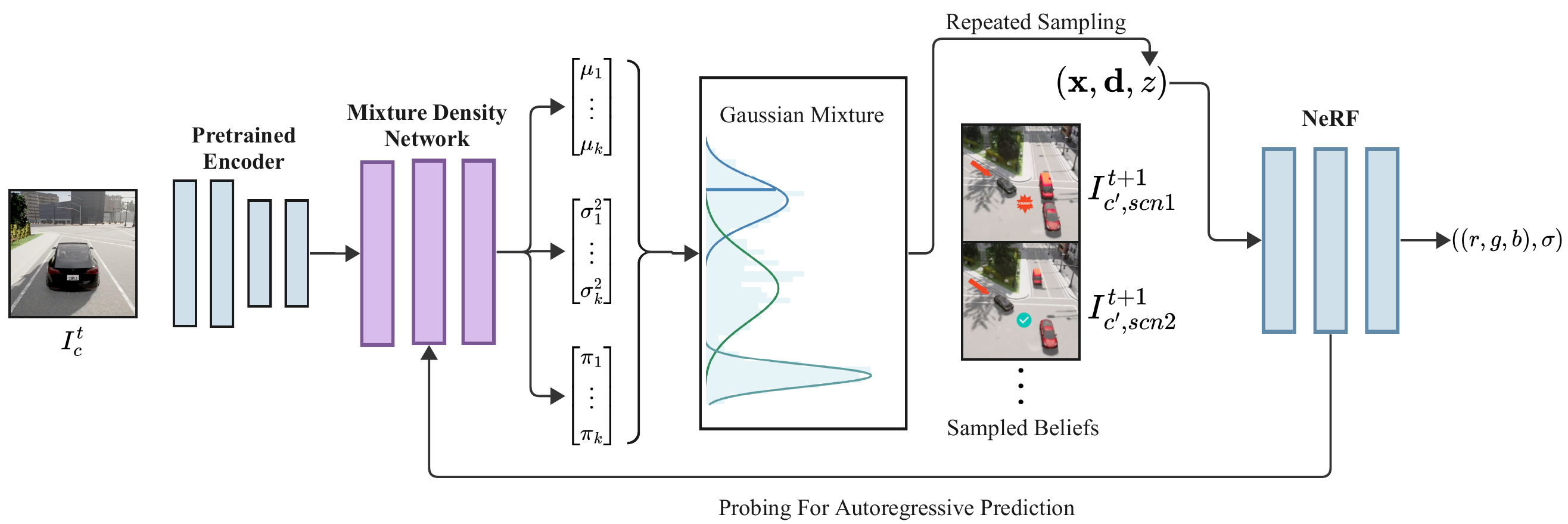}
    \caption{\textbf{Auto-regressive inference in scene prediction}. The input image at timestamp $t$, $I_c^t$, is encoded using the pre-trained encoder from PC-VAE. The corresponding latent distribution is fed into the Mixture Density Network, which predicts a mixture of Gaussians. Each of the $K$ Gaussians is a latent distribution that may correspond to different beliefs at the next timestamp. The mixture of Gaussians is sampled repeatedly for the predicted latent beliefs, visualized as $I_{c',scni}^{t+1}$, representing potentially the $i$th possible outcome. This is used to condition the NeRF to generate 3D views of the scene. To accomplish autoregressive predictions, we probe the NeRF for the location of the car and feed this information back to the pre-trained encoder to predict the scene at the next timestamp.}
    \label{fig:inference}
\end{figure*}

During inference, we observe stochastic behaviors under occlusion, which motivates us to learn a mixture of several Gaussian distributions that potentially denote different scene possibilities.
Therefore, we model the POMDP using a Mixture Density Network (\textit{MDN}), with multi-headed MLPs, that predicts a mixture of $K$ Gaussians.
At any timestamp $t$ the distribution is given as:
\begin{equation}
    q_\phi'(z_t | I_c^{t-1}) = \mog ( q_\phi(z_{t-1} | I_c^ {t-1})) 
\end{equation}
The model is conditioned on the posterior distribution $q_\phi( z_{t-1})$ to learn a predicted posterior distribution $q'_\phi(z_t | I^{t - 1}_c)$ at each timestamp. 
The predicted posterior distribution is given by the mixture of Gaussian:
\begin{equation}
    q_\phi'(z_{t}) = \sum_{i =1}^{K} \pi_i \; \mathcal{N}( \mu_i, \sigma_i^2) 
\end{equation}  
here, $\pi_i$, $\mu_i$, and $\sigma_i^2$ denote the mixture weight, mean, and variance of the $i^{th}$ Gaussian distribution within the posterior distribution. Here, $K$ is the total number of Gaussians. For brevity we remove their conditioning on the posterior $q_\phi (z_{t-1})$ and sampled latent $z_{t-1}$. We sample $z_{t}$ from the mixture of Gaussians $q_\phi'(z_{t})$, where $z_t$ likely falls within one of the Gaussian modes. The configuration corresponding to the mode is reflected in the 3D scene rendered by NeRF.
\paragraph{Loss:}
To optimize the MDN, we minimize a negative log-likelihood function, given by: 
\begin{equation}
    \mathcal{L}_{\textit{MDN}} =
    - \sum_{j=1}^{N}  log \left( \sum_{i=1}^{K} \pi_i \mathcal{N} ( y_j; \mu_i, \sigma_i^2) \right)
\end{equation}
where $y_i \sim q_{\phi} (z_t)$ is sampled from the distribution of latent $z_t$, learned by the encoder, and $N$ denotes the total number of samples.
\paragraph{Inference:}
We consider an unseen ego-centric image and retrieve its posterior $q_\phi (z_t)$ through the encoder.
Next, we predict the possible future posterior distribution $q'_\phi(z_{t+1})$. From the predicted posterior, we sample a scene latent and perform localization. We achieve this via (a) density probing the NeRF or (b) segmenting the rendered novel views using off-the-shelf methods such as YOLO \cite{redmon2016look} (see Fig.~\ref{fig:nvs}). These allow us to retrieve a corresponding Gaussian distribution $q_\phi (z_{t+1})$ in encoder latent space. This is auto-regressively fed back into the MDN to predict the next timestamp. See Fig.~\ref{fig:inference} for an overview of the pipeline.
\section{Results}
\begin{figure}[t]
\begin{floatrow}
\ffigbox[\FBwidth]{\includegraphics[width=.45\textwidth]{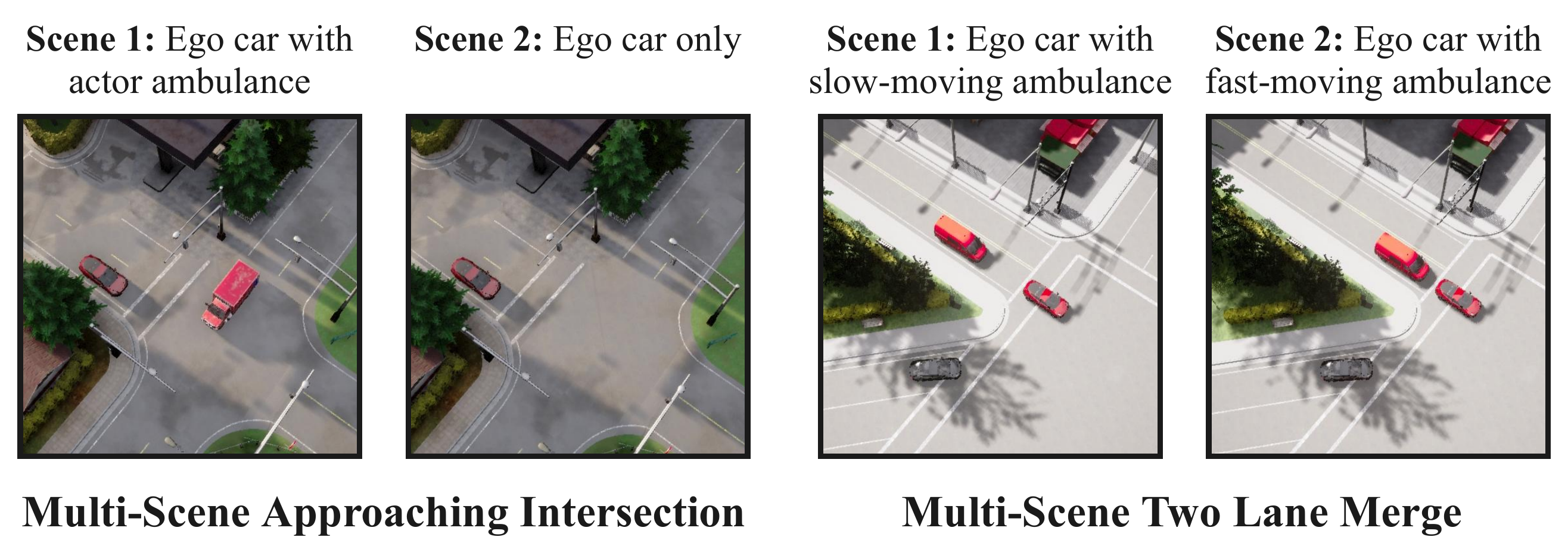}}{\caption{\textbf{Multi-scene CARLA datasets}. Varying car configurations and scenes for the Multi-Scene Two Lane Merge dataset (\textbf{left}) and the Multi-Scene Approaching Intersection dataset (\textbf{right}).}\label{fig:carla_datasets}}
\ffigbox[\FBwidth]{\includegraphics[width=.5\textwidth]{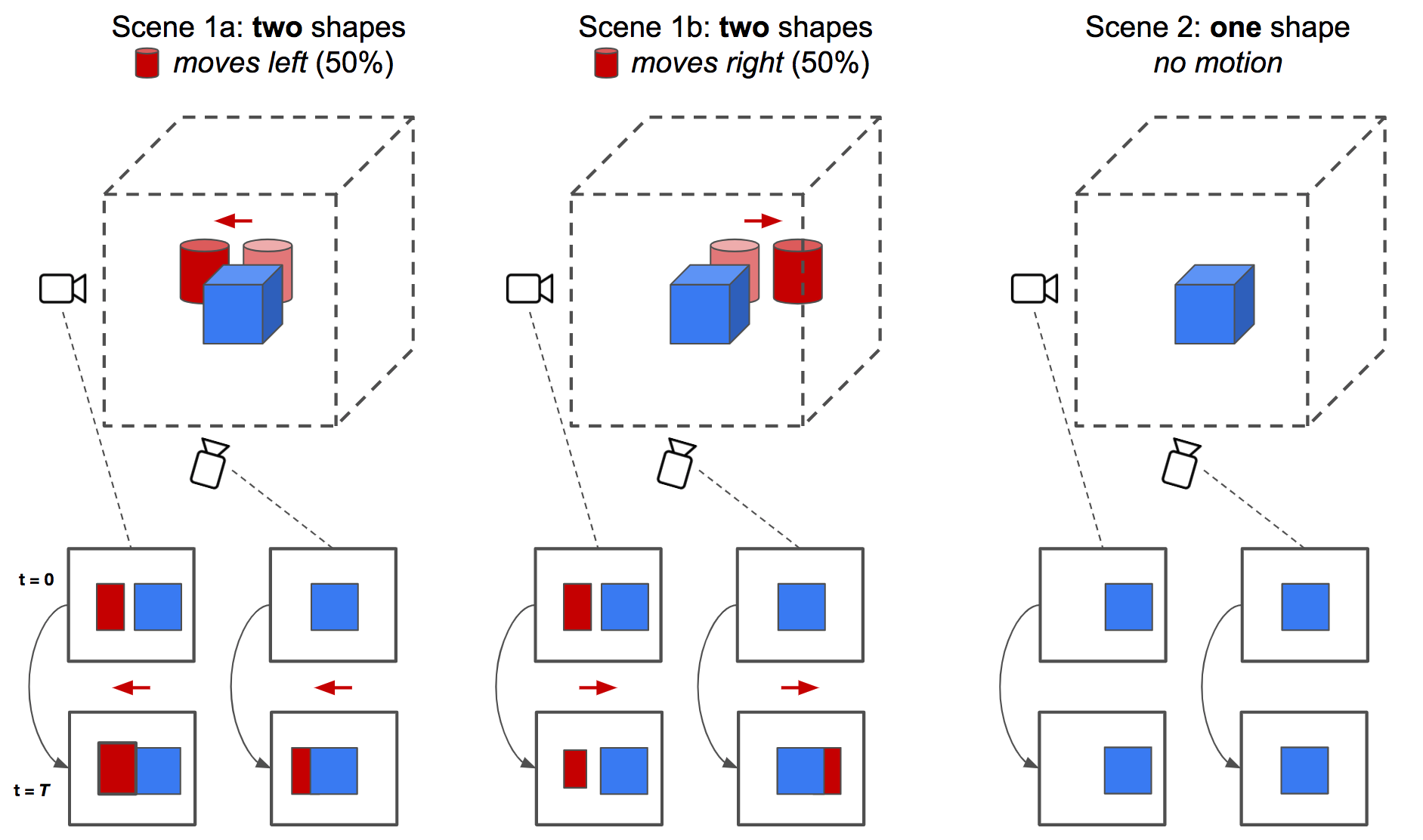}}{\caption{\textbf{Blender dataset}. Blender dataset with a blue cube and a potential red cylinder exhibiting probabilistic temporal movement. The possible occlusions from different camera angles demonstrate how movement needs to be modeled probabilistically.}\label{fig:blender}}
\end{floatrow}
\end{figure}
Decision-making under perceptual uncertainty is a pervasive challenge faced in robotics and autonomous driving, as the real environment is mostly likely partially observable, making it a POMDP.
In a partially observable driving scenario, accurate inference regarding the presence of potentially obscured agents is pivotal.
We evaluate the effectiveness of \oursshort{} on common driving situations with partial observability and added complexity.
We implemented several scenarios in the CARLA driving simulator~\cite{Dosovitskiy17} (see Fig.~\ref{fig:carla_datasets}). A single NVIDIA RTX 3090 GPU is used to train PC-VAE, NeRF, and the MDN. All models, trained sequentially, tend to converge within a combined time frame of 24 hours. A detailed experimental setup can be found in Appendix~\ref{sec:train_details}. We show that, given partially observable 2D inputs, \oursshort{} performs well in predicting latent distributions that represent complete 3D scenes. %
Using these predictions we design a \oursshort-based controller for performing downstream planning tasks. 
\subsection{Data Generation}
\label{sec:data_gen}
We conduct experiments on (a) synthetic blender dataset for principle experiments to test the probabilistic modeling capacities in isolation of the vision encoder (it is visually as simple as possible, but requires the full predictive model proposed in CARFF) and (b) CARLA-based driving datasets for more complex driving scenarios \cite{Dosovitskiy17}. To deliver convincing results, we model these driving scenarios off of related works~\cite{9564424, packer2022is, 10229239, Yu_2019} that concern planning for driving under difficult situations. 
We generate the datasets in 3D by programming an ego object and varying actor objects in different configurations.
\paragraph{Blender synthetic dataset:}
This comprises of a stationary blue cube (ego) accompanied by a red cylinder (actor) that may or may not be present (see Fig.~\ref{fig:blender}).
If the actor is present, it exhibits lateral movement as depicted in Fig.~\ref{fig:blender}. 
This simplistic setting provides an interpretable framework to evaluate our model.
\paragraph{CARLA dataset:} Each dataset is simulated for $N$ timestamps and uses $C=100$ predefined camera poses to capture images of the environment under full observation, partial observation, and no visibility. 
These datasets are modeled after common driving scenarios involving state uncertainty that have been proposed in related works such as Active Visual Planning~\cite{packer2022is}.

\textit{a) Single-Scene Approaching Intersection:} The ego vehicle is positioned at a T-intersection. An actor vehicle traverses the crossing along an evenly spaced, predefined trajectory. We simulate this for $N=10$ timestamps. We mainly use this dataset to predict the evolution of timestamps under full observation.

\textit{b) Multi-Scene Approaching Intersection:} We extend the previous scenario to a more complicated setting with state uncertainty, by making the existence of the actor vehicle probabilistic. A similar intersection crossing is simulated for $N =3$ timestamps for both possibilities. The ego vehicle's view of the actor may be occluded as it approaches the T-intersection over the $N$ timestamps. The ego vehicle either moves forward or halts at the junction (see Fig.~\ref{fig:carla_datasets}).

\textit{c) Multi-Scene Multi-actor Two Lane Merge:} To add more environment dynamics uncertainty, we consider a multi-actor setting at an intersection of two merging lanes. 
We simulate the scenario at an intersection with partial occlusions, with the second approaching actor having variable speed.
Here the ego vehicle can either merge into the left lane before the second actor or after all the actors pass, (see Fig.~\ref{fig:carla_datasets}).
Each branch is simulated for $N=3$ timestamps. 
\subsection{\oursshort{} Evaluation}
\label{sec:carff_eval}
\begin{table*}[h]
    % \vspace*{-\baselineskip}
    \begin{tabular*}{\textwidth}{p{0.12\textwidth}P{0.04\textwidth}P{0.14\textwidth}P{0.14\textwidth}P{0.14\textwidth}P{0.14\textwidth}P{0.11\textwidth}P{0.10\textwidth}P{0.14\textwidth}}
    \toprule
        Method & \shortstack{3D} & \shortstack{Complex  \\ Scenarios} & \shortstack{State \\ Uncertainty} & \shortstack{Dynamics \\ Uncertainty} & Prediction & Planning & \shortstack{Code \\ Released} \\
    \midrule
        \rowcolor{gray!10}
        CARFF & \checkmark & \checkmark & \checkmark & \checkmark & \checkmark & \checkmark & \checkmark \\
        \cite{kosiorek2021nerfvae} & \checkmark & & \checkmark & & & & &  \\
        \rowcolor{gray!10}
        \cite{li20223nerfdy} & \checkmark & \checkmark & & & \checkmark & \checkmark & \\
        \cite{adamkiewicz2022nerfplanning} & \checkmark & \checkmark & & & & \checkmark & \checkmark \\
        \rowcolor{gray!10}
        \cite{packer2022is} & & \checkmark & \checkmark & \checkmark & \checkmark & \checkmark &  \\
    \bottomrule
    \end{tabular*}
    \caption{\textbf{Qualitative comparison of CARFF to related works.} CARFF accomplishes all highlighted objectives as opposed to NeRF-VAE~\cite{kosiorek2021nerfvae}, NeRF for Visuomotor Control~\cite{li20223nerfdy}, Vision-only NeRF Navigation~\cite{adamkiewicz2022nerfplanning}, and AVP~\cite{packer2022is}. We compare whether methods reason in a 3D environment and perform novel view synthesis; work on complex scenarios; predict probabilistically under state and dynamics uncertainty; forecast into the future; and use model predictions for decision-making.}
    \label{tab:related_work}
\end{table*}
A desirable behavior from our model is that it should predict a complete set of possible scenes consistent with the given ego-centric image, which could be partially observable.
This is crucial for autonomous driving in unpredictable environments as it ensures strategic decision-making based on potential hazards.
To achieve this we require a rich PC-VAE latent space, high-quality novel view synthesis, and auto-regressive probabilistic predictions of latents at future timestamps.
We evaluate \oursshort{} on a simple synthetic blender-based dataset and each CARLA-based dataset. Additionally, we extend our model application to a hand-manipulation dataset in Appendix~\ref{sec:appendix-datasets}.

\paragraph{Comparisons with related work:}
We attempt to compare CARFF to existing approaches. 
NeRF-VAE has the most comparable objective, but during our experiments, it collapse to black using CARLA datasets.
We make further qualitative comparisons to other most similar methods in Tab.~\ref{tab:related_work}, but none aligns with ours enough to make any possible quantitative comparisons.
\paragraph{Evaluation on blender dataset:}
In Fig.~\ref{fig:blender}, for both Scene 1a and 1b, our model correctly forecasts the lateral movement of the cylinder to be in either position approximately 50\% of the time, considering a left viewing angle. In Scene 2, with the absence of the red cylinder in the input camera angle, the model predicts the potential existence of the red cylinder approximately 50\% of the time, and predicts lateral movements with roughly equal probability. This validates PC-VAE's ability to predict and infer occlusions in the latent space, aligning with human intuitions. These intuitions, shown in the Blender dataset's simple scenes, can transfer to driving scenarios in our CARLA datasets.
\begin{figure*}
    \centering
    \includegraphics[width= 0.8 \textwidth]{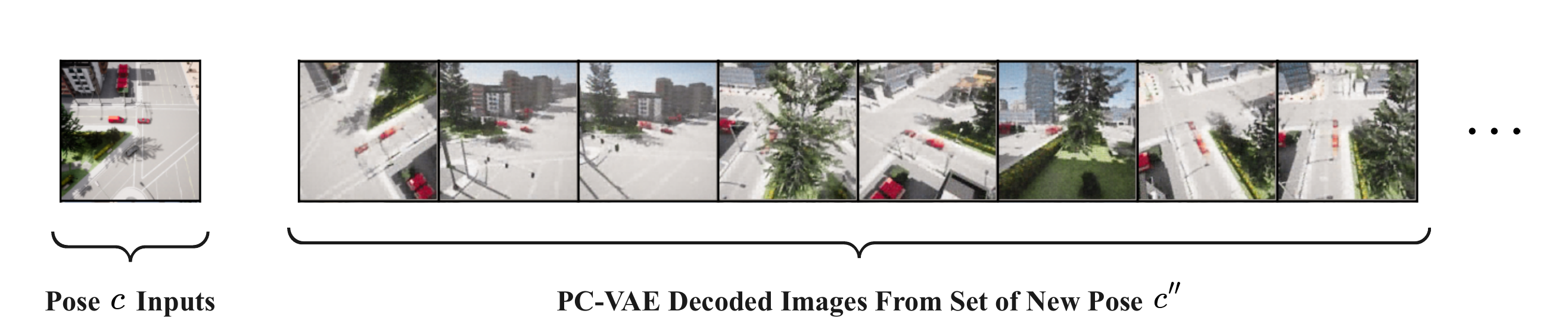}
    \caption{\textbf{PC-VAE reconstructions}. The encoder input, $I^t_c$, among the other ground truth images $I_c$ viewed from camera pose $c$ at different timestamps, is reconstructed across a new set of poses $c''$ respecting timestamp $t$, generating $I^t_{c''}$. A complete grid is in Appendix \ref{sec:appendix-vae}.}
    \label{fig:recons}
\end{figure*}
\paragraph{PC-VAE performance and ablations:}
We evaluate the performance of PC-VAE on CARLA datasets with multiple encoder architectures. We show that PC-VAE effectively reconstructs complex environments involving variable scenes, actor configurations, and environmental noise given potentially partially observable inputs (see Fig.~\ref{fig:recons}). We calculated an average Peak Signal-to-Noise Ratio (PSNR) over the training data, as well as novel view encoder inputs. 
To evaluate the quality of the latent space generated by the encoder, we utilize t-SNE \cite{vanDerMaaten2008} plots to visualize the distribution of latent samples for each image in a given dataset (see Appendix~\ref{sec:appendix-vae}). 
We introduce a Support Vector Machine (SVM)~\cite{hearst1998support} based metric to measure the visualized clustering quantitatively, where a higher value indicates better clustering based on timestamps.  
Most latent scene samples are separable by timestamps, which indicates that the latents are view-invariant.
Samples that are misclassified or lie on the boundary usually represent partially or fully occluded regions.
This is desirable for forecasting, as it enables us to model probabilistic behavior over these samples.
In this process, balancing KL divergence weight scheduling maintains the quality of the PC-VAE's latent space and reconstructions (see Appendix~\ref{sec:train_details}).
Additionally, we substantiate the benefits of our PC-VAE encoder architecture through our ablations (see Appendix~\ref{sec:appendix-ablations}).
\begin{table*}[t]
% \vspace*{-\baselineskip}
\begin{floatrow}
\capbtabbox{%
  \begin{tabular*}{0.45\textwidth}{p{0.19\textwidth}P{0.12\textwidth}P{0.12\textwidth}}
        \toprule
        \mbox{Ground Truth} \mbox{Prediction} Pair & Avg. PSNR (Scene 1) & Avg. PSNR (Scene 2) \\
        \midrule
        \multicolumn{3}{p{0.4\textwidth}}{\textbf{Single-Scene Intersection}} \\
        \midrule
        Matching Pairs & \textbf{29.06} & N.A\\
        Un-matching P. & 24.01 & N.A \\
        \midrule
        \multicolumn{3}{p{0.4\textwidth}}{\textbf{Multi-Scene Intersection}} \\
        \midrule
        Matching Pairs & \textbf{28.00} & \textbf{28.26} \\
        Un-matching P. & 23.27 & 24.56 \\
        \midrule
        \multicolumn{3}{p{0.4\textwidth}}{\textbf{Multi-Scene Two Lane Merge}} \\
        \midrule
        Matching Pairs & \textbf{28.14} & \textbf{28.17} \\
        Un-matching P. & 22.74 & 23.32\\
        \bottomrule
    \end{tabular*}
}{
    \caption{\textbf{Averaged PSNR for fully observable 3D predictions}. \oursshort{} correctly predicts scene evolution across all timestamps for each dataset. The average PSNR is high for predictions $\hat{I_{t_i}}$ and matching ground truths, $I_{t_i}$. PSNR values for incorrect correspondences, $\hat{I_{t_i}}, I_{t_j}$, is a result of matching surroundings. See complete table in Appendix~\ref{sec:appendix-vae}.}
    \label{tab:predictions}
}%
\capbtabbox{%
    \begin{tabular*}{0.45\textwidth}{p{0.25\textwidth}>{\raggedleft\arraybackslash}p{0.08\textwidth}>{\raggedleft\arraybackslash}p{0.11\textwidth}}
        \toprule
        \multicolumn{3}{p{0.45\textwidth}}{\textbf{Multi-Scene Intersection}} \\
        Controller Type & Actor & No Actor \\
        \midrule
        Underconfident & $30/30$ & $0/30$ \\
        Overconfident & $0/30$ & $30/30$ \\
        \oursshort{} ($n$=2) & $17/30$ & $30/30$ \\
        \textbf{\oursshort{} ($\vect{n}$=10)} & $\mathbf{30/30}$ & $\mathbf{30/30}$ \\
        \oursshort{} ($n$=35) & $30/30$ & $19/30$ \\
        \toprule
        \multicolumn{3}{p{0.5\textwidth}}{\textbf{Multi-Scene Two Lane Merge}} \\
        Controller Type & Fast & Slow \\
        \midrule
        Underconfident & $30/30$ & $0/30$ \\
        Overconfident & $0/30$ & $30/30$ \\
        \oursshort{} ($n$=2) & $21/30$ & $30/30$ \\
        \textbf{\oursshort{} ($\vect{n}$=10)} & $\mathbf{30/30}$ & $\mathbf{30/30}$ \\
        \oursshort{} ($n$=35) & $30/30$ & $22/30$ \\
        \bottomrule
    \end{tabular*}
}{
  \caption{\textbf{Planning in 3D with controllers with varying sampling numbers $\vect{n}$}. CARFF-based controllers outperform baselines in success rate over 30 trials. For $n$ = 10, the CARFF-based controller consistently chooses the optimal action in potential collision scenarios.To maintain consistency, we use one single image input across 30 trials.}
  \label{tab:planning}
}
\end{floatrow}
\end{table*}

\paragraph{3D novel view synthesis:}
Given an unseen ego-centric view with potentially partial observations, our method maintains all possible current state beliefs in 3D, and faithfully reconstructs novel views from arbitrary camera angles for each belief. Fig.~\ref{fig:nvs} illustrates one of the possible 3D beliefs that \oursshort{} holds. 
This demonstrates our method's ability to generate 3D beliefs that could be used for novel view synthesis in a view-consistent manner. 
Our model's ability to achieve accurate and complete 3D environmental understanding is important for applications like prediction-based planning.
\paragraph{Inference under full and partial observations:}
Under full observation, we use MDN to predict the subsequent car positions in all three datasets. PSNR values are calculated based on bird-eye view NeRF renderings and ground truth bird-eye view images of the scene across different timestamps. In Tab.~\ref{tab:predictions} we report the PSNR values for rendered images over the predicted posterior with the ground truth images at each timestamp. 
We also evaluate the efficacy of our prediction model using the accuracy curve given in Fig.~\ref{fig:curves}. 
This represents \oursshort{}'s ability to generate stable beliefs, without producing incorrect predictions, based on actor(s) localization results. For each number of samples between $n = 0$ to $n = 50$, we choose a random subset of $3$ fully observable ego images and take an average of the accuracies.
In scenarios with partial observable ego-centric images where several plausible scenarios exist, we utilize recall instead of accuracy using a similar setup.  
This lets us evaluate the encoder's ability to avoid false negative predictions of potential danger. 

Fig.~\ref{fig:curves} shows that our model achieves high accuracy and recall in both datasets, demonstrating the ability to model state uncertainty (Approaching Intersection) and dynamic uncertainty (Two Lane Merge).
The results indicate \oursshort's resilience against randomness in resampling, and completeness in probabilistic modeling of the belief space. Given these observations, we now build a reliable controller to plan and navigate through complex scenarios.
\subsection{Planning}
In all our experiments, the ego vehicle must make decisions to advance under certain observability. The scenarios are designed such that the ego views contain partial occlusion and the state of the actor(s) is uncertain in some scenarios.

In order to facilitate decision-making using \oursshort, we design a controller that takes ego-centric input images and outputs an action. 
Decisions are made incorporating sample consistency from the mixture density network. For instance, the controller infers occlusion and promotes the ego car to pause when scenes alternate between actor presence and absence in the samples. We use the two multi-scene datasets to assess the performance of the \oursshort-based controller as they contain actors with potentially unknown behaviors.

\begin{figure*}
% \vspace*{-\baselineskip}
\begin{subfigure}{0.27\textwidth}
        \centering
        \resizebox{\textwidth}{!}
        {\begin{tikzpicture}

\definecolor{darkslategray38}{RGB}{38,38,38}
\definecolor{indianred}{RGB}{205,92,92}
\definecolor{lightgray204}{RGB}{204,204,204}
\definecolor{steelblue}{RGB}{70,130,180}
\Large 

\begin{axis}[
axis line style={lightgray204},
tick align=outside,
axis line style={lightgray204},
legend cell align={left},
legend style={
  fill opacity=0.8,
  draw opacity=1,
  text opacity=1,
  at={(0.97,0.03)},
  anchor=south east,
  draw=lightgray204,
},
x grid style={lightgray204},
xlabel=\textcolor{darkslategray38}{\LARGE{Num Samples}},
xmajorgrids,
xmin=0, xmax=50.1,
xtick style={color=darkslategray38},
y grid style={lightgray204},
ylabel=\textcolor{darkslategray38}{\%},
ymajorgrids,
ymin=0.4, ymax=1.02,
ytick pos=left,
ytick style={color=darkslategray38}
]

\addlegendentry{Accuracy}
\addplot [line width=4pt, indianred]
table {%
1 1.0
2 1.0
3 1.0
4 1.0
5 1.0
6 1.0
7 1.0
8 1.0
9 1.0
10 1.0
11 1.0
12 1.0
13 1.0
14 1.0
15 1.0
16 1.0
17 1.0
18 1.0
19 1.0
20 1.0
21 1.0
22 0.99
23 1.0
24 0.9867
25 1.0
26 0.9815
27 0.9880
28 0.9843
29 0.9810
30 0.98
31 0.9732
32 0.9788
33 0.9797
34 0.9706
35 0.9714
36 0.9722
37 0.9730
38 0.9737
39 0.9787
40 0.975
41 0.9712
42 0.9648
43 0.9730
44 0.9721
45 0.9711
46 0.9758
47 0.9732
48 0.975
49 0.9780
50 0.974
};
\addlegendentry{Recall}

\addplot [line width=4pt, dashed, steelblue]
table {%
0 0
1 0.5
2 0.5
3 0.5
4 1.0
5 1.0
6 1.0
7 1.0
8 1.0
9 1.0
10 1.0
11 1.0
12 1.0
13 1.0
14 1.0
15 1.0
16 1.0
17 1.0
18 1.0
19 1.0
20 1.0
21 1.0
22 1.0
23 1.0
24 1.0
25 1.0
26 1.0
27 1.0
28 1.0
29 1.0
30 1.0
31 1.0
32 1.0
33 1.0
34 1.0
35 1.0
36 1.0
37 1.0
38 1.0
39 1.0
40 1.0
41 1.0
42 1.0
43 1.0
44 1.0
45 1.0
46 1.0
47 1.0
48 1.0
49 1.0
50 1.0
};

\end{axis}

\end{tikzpicture}}
        \caption{Intersection}
    \end{subfigure}%
    \begin{subfigure}{0.27\textwidth}
        \centering
        \resizebox{\textwidth}{!}{\begin{tikzpicture}

\definecolor{darkslategray38}{RGB}{38,38,38}
\definecolor{indianred}{RGB}{205,92,92}
\definecolor{lightgray204}{RGB}{204,204,204}
\definecolor{steelblue}{RGB}{70,130,180}
\Large 

\begin{axis}[
axis line style={lightgray204},
tick align=outside,
axis line style={lightgray204},
legend cell align={left},
legend style={
  fill opacity=0.8,
  draw opacity=1,
  text opacity=1,
  at={(0.97,0.03)},
  anchor=south east,
  draw=lightgray204,
},
x grid style={lightgray204},
xlabel=\textcolor{darkslategray38}{\LARGE{Num Samples}},
xmajorgrids,
xmin=0, xmax=50.1,
xtick style={color=darkslategray38},
y grid style={lightgray204},
ylabel=\textcolor{darkslategray38}{\%},
ymajorgrids,
ymin=0.4, ymax=1.02,
ytick pos=left,
ytick style={color=darkslategray38}
]

\addlegendentry{Accuracy}
\addplot [line width=4pt, indianred]
table {%
1 1.0
2 1.0
3 1.0
4 1.0
5 1.0
6 1.0
7 1.0
8 1.0
9 1.0
10 1.0
11 1.0
12 1.0
13 1.0
14 1.0
15 1.0
16 1.0
17 1.0
18 1.0
19 1.0
20 1.0
21 1.0
22 1.0
23 1.0
24 1.0
25 1.0
26 1.0
27 1.0
28 1.0
29 0.99
30 1.0
31 0.9932
32 0.9925
33 0.981
34 0.98
35 0.9814
36 0.9856
37 0.9830
38 0.9771
39 0.9764
40 0.98
41 0.9752
42 0.9778
43 0.9840
44 0.9791
45 0.9771
46 0.9768
47 0.9834
48 0.9797
49 0.9771
50 0.98
};

\addlegendentry{Recall}
\addplot [line width=4pt, dashed, steelblue]
table {%
0 0
1 0.5
2 0.5
3 1.0
4 1.0
5 1.0
6 1.0
7 1.0
8 1.0
9 1.0
10 1.0
11 1.0
12 1.0
13 1.0
14 1.0
15 1.0
16 1.0
17 1.0
18 1.0
19 1.0
20 1.0
21 1.0
22 1.0
23 1.0
24 1.0
25 1.0
26 1.0
27 1.0
28 1.0
29 1.0
30 1.0
31 1.0
32 1.0
33 1.0
34 1.0
35 1.0
36 1.0
37 1.0
38 1.0
39 1.0
40 1.0
41 1.0
42 1.0
43 1.0
44 1.0
45 1.0
46 1.0
47 1.0
48 1.0
49 1.0
};

\end{axis}

\end{tikzpicture}}
        \caption{Lane Merge}
    \end{subfigure}
    \caption{\textbf{Multi-Scene dataset accuracy and recall curves from predicted beliefs.} We test our framework across $n = 1$ and $n = 50$ samples from MDN's predicted latent distributions from ego-centric image input. Across the number of samples $n$, we achieve an ideal margin of belief state coverage generated under partial observation (recall), and the proportion of correct beliefs sampled under full observation (accuracy). As we significantly increase the number of samples, the accuracy starts to decrease due to randomness in latent distribution resampling.}
    \label{fig:curves}
\end{figure*}
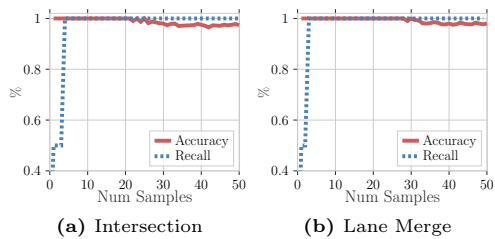

To design an effective controller, we need to find a balance between accuracy and recall (see Fig.~\ref{fig:curves}). 
A lowered accuracy from excessive sampling means unwanted randomness in the predicted state. However, taking insufficient samples would generate low recall i.e., not recovering all plausible states.
This would lead to incorrect predictions as we would be unable to account for the plausible uncertainty present in the environment. 
To achieve optimal balance, we designed an open-loop planning controller using a sampling strategy that generates $n = 2, 10, 35$ samples. The hyperparameter $n$ is tuned per scene for peak performance but is expected to remain relatively stable across scenes. We demonstrate that $n = 10$ performs well consistently in varying CARLA scenarios (Fig.~\ref{fig:curves}) and do not anticipate this being very different for other experiments.

For sampling values that lie on the borders of the accuracy and recall margin, for example, $n = 2$ and $35$, we see that the \oursshort-based controller obtains lower success rates, whereas $n = 10$ produces the best result. For actor exists and fast-actor scenes in Tab.~\ref{tab:planning}, we consider occluded ego-centric inputs to test the controller's ability to avoid collisions. For no-actor and slow-actor scenes, we consider state observability and test the controllers' ability to recognize the optimal action to advance. Across the two datasets, the overconfident controller will inevitably experience collisions in case of a truck approaching, since it does not cautiously account for occlusions. On the other hand, an overly cautious approach results in stasis, inhibiting the controller's ability to advance in the scene. \
This nuanced decision-making using \oursshort-based controller is especially crucial in driving scenarios, as it enhances safety and efficiency by adapting to complex and unpredictable road environments, thereby fostering a more reliable and human-like response in autonomous vehicles.
\section{Discussion}
\label{sec:discussion}
\paragraph{Limitations:}
Like other NeRF-based methods, \oursshort{} currently relies on posed images of specific scenes such as road intersections, limiting its direct applicability to unseen environments. However, we anticipate enhanced generalizability with the increasing deployment of cameras around populated areas, such as traffic cameras at intersections. Additionally, handling very complex dynamics with an extremely large number of actors still poses a challenge for our method, requiring per-scene optimization to balance comprehensive dynamics modeling against accuracy. Potentially stronger models in the near future may offer a promising avenue for further enhancements in this regard.
% \vspace*{-\baselineskip}
\paragraph{Conclusion:}
We present \oursshort{}, a novel method for probabilistic 3D scene forecasting from partial observations. By employing a Pose-Conditional VAE, a NeRF conditioned on the learned posterior, and a mixture density network that forecasts future scenes, we effectively model, predict, and plan in complex environments with state and dynamics uncertainty in a POMDP. 
We further demonstrate the capabilities of our method in simulated autonomous driving scenarios.
Overall, \oursshort{} offers an intuitive framework to perceiving, forecasting, and acting under uncertainty that could prove invaluable for vision-based algorithms in unstructured environments.

\definecolor{regulargreen}{rgb}{0.0, 0.5, 0.0}
\newcolumntype{P}[1]{>{\centering\arraybackslash}p{#1}}
\newcommand{\minus}{\scalebox{0.75}[1.0]{$-$}}
\newcommand{\xmark}{\ding{55}}
\appendix
\definecolor{bluegray}{rgb}{0.27, 0.42, 0.81}

\clearpage

\section{Datasets}
\label{sec:appendix-datasets}

\subsection{CARLA Datasets} 
A complete figure of the actor and ego configurations across scenes and the progression of timestamps for the Single-Scene Approaching Intersection, Multi-Scene Approaching Intersection, and the Multi-Scene Two Lane Merge is visualized in Fig.~\ref{fig:datasets}.

\subsection{Hand-manipulation Dataset} 
\begin{sidewaysfigure*}
    \centering
    \includegraphics[width= \textwidth]{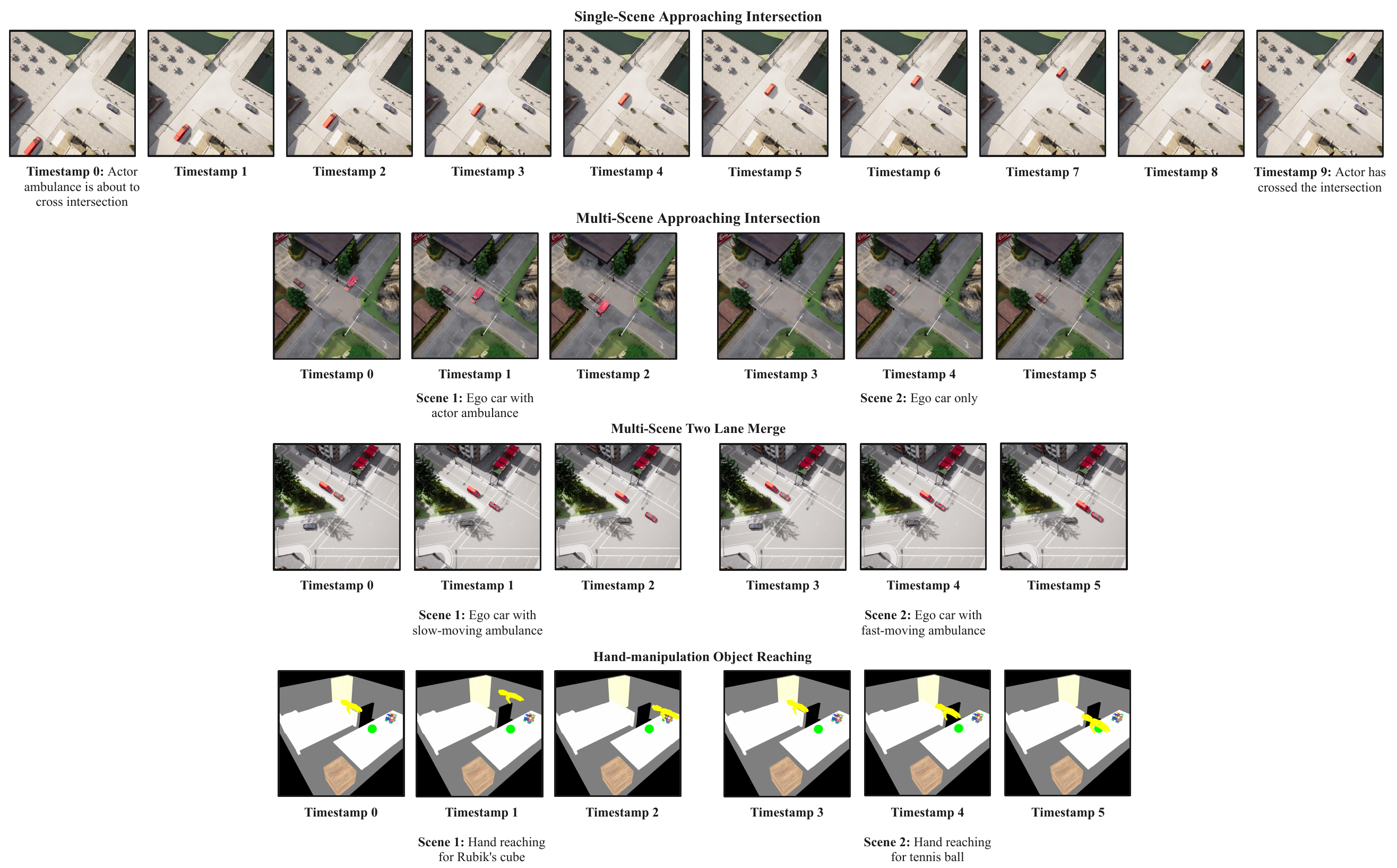}
    \caption{\textbf{Single-Scene Approaching Intersection, Multi-Scene Approaching Intersection, Multi-Scene Two Lane Merge, and Hand-manipulation Object Reaching Datasets}. The actor and ego car configurations and the hand configurations for the timestamps and scenes of the three CARLA datasets, and the hand-manipulation dataset are visualized at a single camera pose. The colors of the cars for the Multi-Scene Approaching Intersection have been slightly modified for greater contrast and visual clarity in the paper. }
    \label{fig:datasets}
\end{sidewaysfigure*}

We generate an additional hand-manipulation dataset involving a robotic manipulator engaged in probabilistic reaching tasks utilizing VUER~fig\cite{vuer}. The experimental setup consists of two target objects placed on a table in a room: a Rubik's cube and a green tennis ball. The robotic hand's configurations during the reaching and grasping phases for both objects are illustrated in Fig.~\ref{fig:datasets}.

\section{Implementation Details}
\label{sec:train_details}

\begin{figure*}
    \centering
    \includegraphics[width=\textwidth]{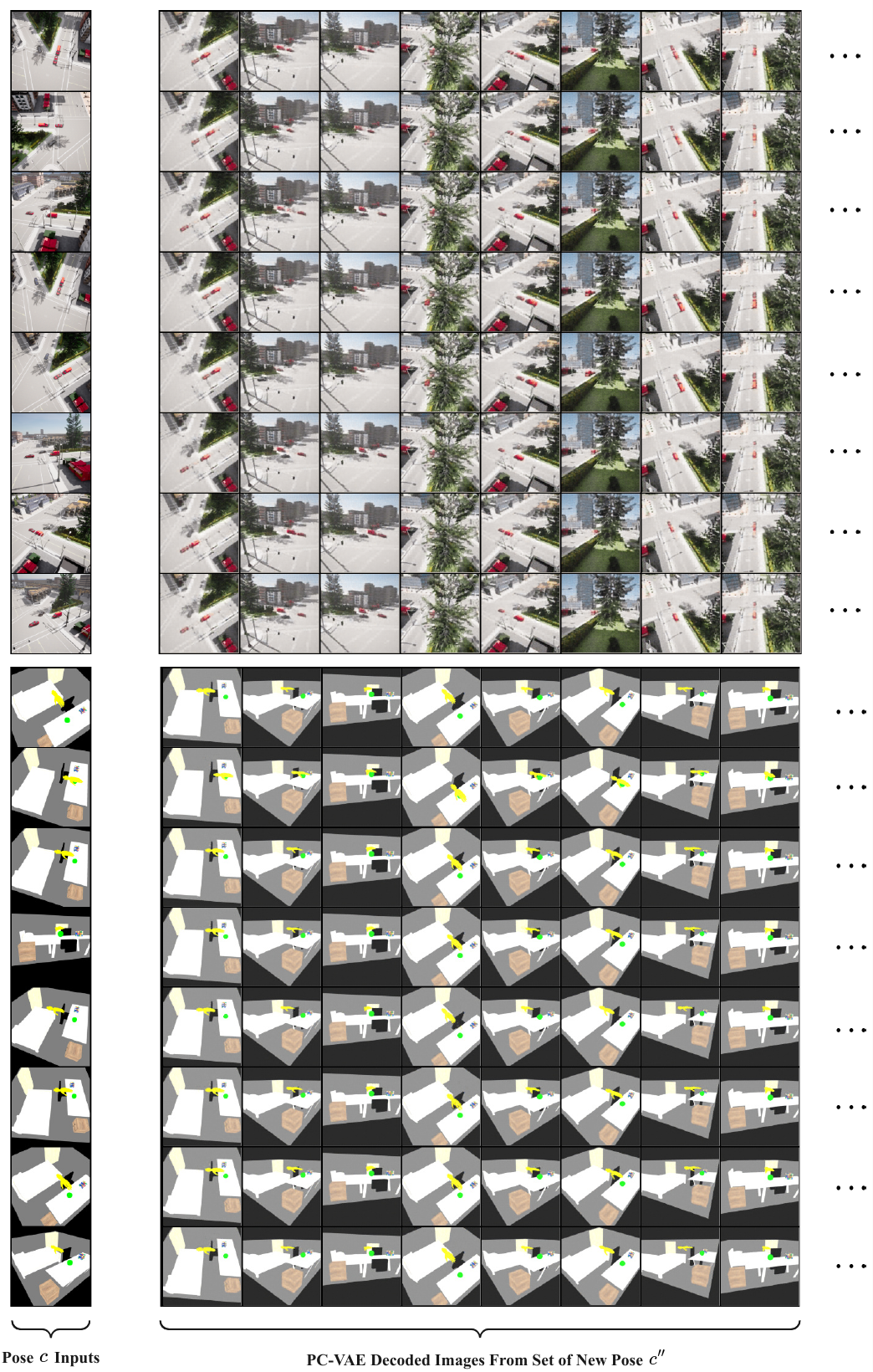}
    \caption{\textbf{PC-VAE encoder inputs, ground truth timestamps, and reconstructions for a CARLA dataset and Hand-manipulation dataset}. The encoder input, $I^t_c$, among the other ground truth images $I_c$ viewed from camera pose $c$ at different timestamps, is reconstructed across a new set of poses $c''$ respecting timestamp $t$, generating $I^t_{c''}$. This is a full grid of the reconstructions.}
    \label{fig:recons}
\end{figure*}

\subsection{Pose-Conditional VAE}

\begin{table}[h]
    \centering
    \begin{tabular*}{0.6\textwidth}{l>{\raggedleft\arraybackslash}p{0.32\textwidth}}
    \toprule
        \multicolumn{2}{p{0.6\textwidth}}{\textbf{PC-VAE Hyperparameters}} \\
    \midrule
        Latent Size & 8 \\
        LR & $0.004$ \\
        KLD Weight Start & $0.000001$ \\
        KLD Weight End & $0.00001-0.00004$* \\
        KLD Increment Start & $50$ epochs \\
        KLD Increment End & $80$ epochs \\
    \bottomrule
    \end{tabular*}
    \caption{\textbf{PC-VAE experimental setup and hyperparameters.} The main hyperparameters in PC-VAE training on the three datasets are latent size, LR, and KLD weight. For KLD scheduling, the KLD increment start refers to the number of epochs at which the KLD weight begins to increase from the initial KLD weight. KLD increment end is the number of epochs at which the KLD weight stops increasing at the maximum KLD weight. The asterisk (*) marks the hyperparameter that is dataset-dependent.}
    \label{tab:hyperparams}
\end{table}

\paragraph{Architecture:}

We implement PC-VAE on top of a standard PyTorch VAE framework.
The encoder with convolutional layers is replaced with a single convolutional layer and a Vision Transformer (ViT) Large 16~\cite{dosovitskiy2021image} pre-trained on ImageNet~\cite{russakovsky2015imagenet}. 
We modify fully connected layers to project ViT output of size $1000$ to mean and variances with size of the latent dimension, 8.
During training, the data loader returns the pose of the camera angle represented by an integer value. This value is one-hot encoded and concatenated to the re-parameterized encoder outputs, before being passed to the decoder. The decoder input size is increased to add the number of poses to accommodate the additional pose information.

\paragraph{Optimization:}
We utilize a single RTX 3090 graphics card for all our experiments.
The PC-VAE model takes approximately $22$ hours to converge using this GPU.
During this phase, we tune various hyperparameters including the latent size, learning rate and KL divergence loss weight to establish optimal training tailored to our model (see Tab.~\ref{tab:hyperparams}). 
In order to optimize for the varied actor configurations and scenarios generated within the CARLA~\cite{Dosovitskiy17} simulator, we slightly adjust hyperparameters differently for each dataset.

The learning rate (LR) and KL divergence (KLD) weight are adjusted to find an appropriate balance between the effective reconstruction of pose conditioning in the latent space, and the regularization of latents. Regularization pushes the latents toward Gaussian distributions and keeps the non-expressive latents in an over-parameterized latent space to be standard normal. This stabilizes the sampling process and ensures stochastic behavior of latent samples in case of occlusion.
To achieve this balance, we use a linear KLD weight scheduler, where the weight is initialized at a low value for KLD increment start epoch (see Tab.~\ref{tab:hyperparams}). 
This allows the model to initially focus on achieving highly accurate conditioned reconstructions. The KLD weight is then steadily increased until KLD increment end epoch is reached, ensuring probabilistic behavior under partial observability.

\subsection{Mixture Density Network}

With the POMDP, the mixture density network (MDN) takes in the mean and variances of the latent distributions of the current belief state $  q_\phi(z_{t-1} | I_c^ {t-1}) $ and outputs the next belief state's estimated posterior distribution. To better model the uncertainty of the predicted belief state distribution, the output is a mixture of Gaussian $q_\phi'(z_t | I_c^{t-1})$ modeled through a multi-headed MLP.

\paragraph{Architecture:} 
The shared backbone simply contains $2$ fully connected layers and rectified linear units (ReLU) activation with hidden layer size of $512$. 
Additional heads with $2$ fully connected layers are used to generate $\mu_i$ and $\sigma_i^2$. The mixture weight, $\pi_i$, is generated from a $3$ layer MLP network. We limit the number of Gaussians, $K = 2$.  

\paragraph{Optimization:}
We train our network for $30,000$ epochs using the batch size of $128$ and an initial LR of $0.005$, and apply LR decay to optimize training. This takes approximately $30$ minutes to train utilizing the GPU. 
During training, the dataloader outputs the means and variances at the current timestamp and indexed view, and the means and variances for the next timestamp, at a randomly sampled neighboring view. This allows the MDN to learn how occluded views advance into all the possible configurations from potentially unoccluded neighboring views, as a mixture of Gaussian. 

At each iteration, the negative log-likelihood loss is computed for $1000$ samples drawn from the predicted mixture of distributions $q_\phi'(z_t | I_c^{t-1})$ with respect to the ground truth distribution $q_\phi(z_t | I_c^{t})$. 
While the MDN is training, additional Gaussian noise, given by $\epsilon \sim \mathcal{N}(0, \sigma^2)$, is added to the means and variances of the current timestamp $t - 1$, where $\sigma \in [0.001, 0.01]$. The Gaussian noise and LR decay help prevent overfitting and reduce model sensitivity to environmental artifacts like moving trees, moving water, etc.

\subsection{NeRF}

\paragraph{Architecture: }
We implement our NeRF as a decoder to our belief state to recover 3D observations utilizing an existing PyTorch implementation of Instant-NGP~\cite{muller2022instant}. 
We concatenate the latents to the inputs of two parts of the Instant-NGP architecture: the volume density network, $\sigma(\mathbf{x})$, for the density values, and the color network, $C(\mathbf{r})$, for conditional RGB generation. 
While the overall architecture is kept constant, the input dimensions of each network are modified to allow additional latent concatenation.
\paragraph{Optimization: }
Empirically, we observe that it is essential to train the NeRF such that it learns the distribution of scenes within the PC-VAE latent space.
Using only pre-defined learned samples to train may run the risk of relying on non-representative samples.
On the other hand, direct re-sampling during each training iteration in Instant-NGP may lead to delayed training progress, due to NeRF's sensitive optimization. 
%
% To attempt to address these issues, we construct a mixed training procedure. 
In our optimization procedure, we use an LR of $0.002$ along with an LR decay and start with pre-defined latent samples. Then we slowly introduce the re-sampled latents.
We believe that this strategy progressively diminishes the influence of a single sample, while maintaining efficient training. 
Based on our observations, this strategy contributes towards Instant-NGP's ability to rapidly assimilate fundamental conditioning and environmental reconstruction, while simultaneously pushing the learning process to be less skewed towards a single latent sample.
\section{GUI Interface}
\begin{figure*}[h]
    \centering
    \includegraphics[width= 0.6\textwidth]{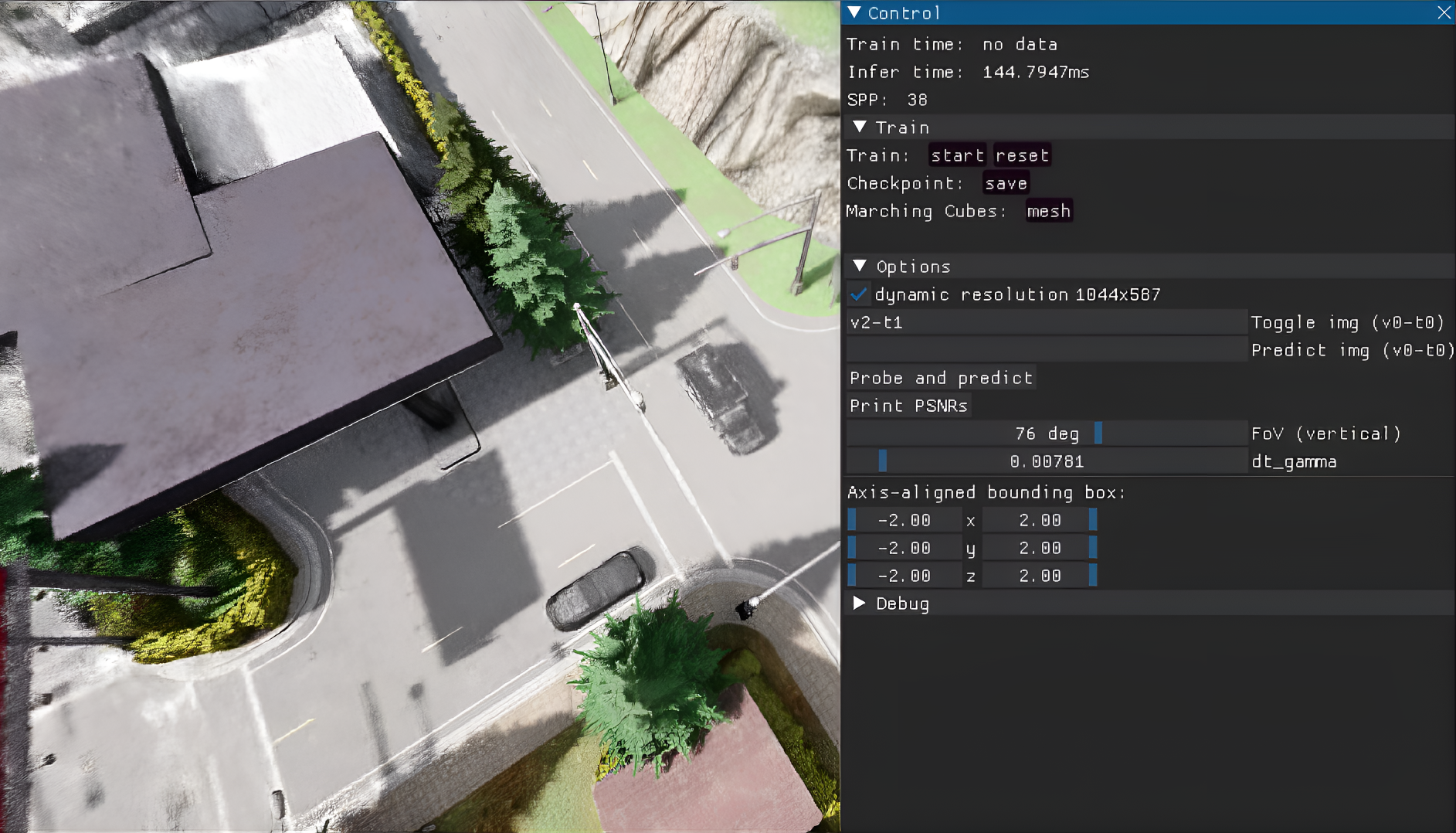}
    \caption{\textbf{NeRF graphical user interface.} The GUI allows us to toggle and predict with an input image path. The probe and predict function probes the current location of the car and predicts the next. The screenshot is sharpened for visual clarity in the paper.}
    \label{fig:nerfgui}
\end{figure*}

\begin{figure}[h]
    \centering
    \includegraphics[width=0.8\textwidth]{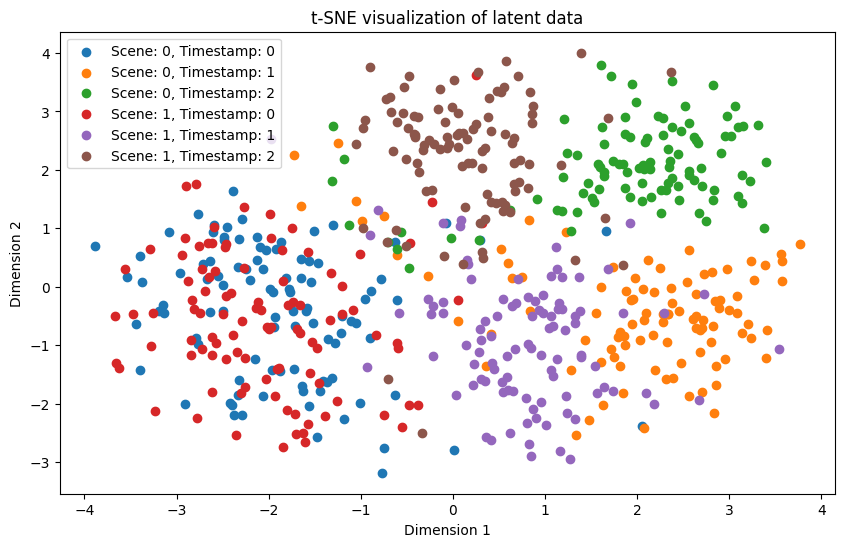}
    \caption{\textbf{Latent sample distribution clustering}. The distributions of latent samples for the Multi-Scene Two Lane Merge dataset are separable through t-SNE clustering. In the figure, the clusters for \textit{Scene 0, Timestamp 0} and \textit{Scene 1, Timestamp 0} overlap in distribution because they represent the same initial state of the environment under dynamics uncertainty.}
    \label{fig:clustering}
\end{figure}
For ease of interaction with our inference pipeline, our NeRF loads a pre-trained MDN checkpoint, and we build a graphical user interface (GUI) using DearPyGUi for visualization purposes.
We implement three features in the GUI: (a) predict, (b) probe and predict, and (c) toggle.

\paragraph{Predict:}
% \textit{Predict:}
We implement the function to perform prediction directly from a given image path in the GUI. We use the distribution $q_{\phi}(z_{t - 1}|I^{t - 1}_c)$ from PC-VAE encoder, corresponding to the input image $I^{t - 1}_c$, to predict the latent distribution for the next timestamp belief state $q_{\phi}'(z_t|I^{t - 1}_c)$. This process is done on the fly through the MDN. 
A sample from the predicted distribution is then generated and used to condition the NeRF. This advances the entire scene to the next timestamp. 

\paragraph{Probe and predict:}
The sampled latent from the predicted distribution does not correspond to a singular distribution and hence we can not directly predict the next timestamp. 
To make our model auto-regressive in nature, we perform density probing. 
We probe the density of the NeRF at the possible location coordinates of the car to obtain the current timestamp and scene.
This is then used to know the actual state sampled from the belief state probability distributions.
The new distribution enables auto-regressive predictions using the predict function described above.

\paragraph{Toggle:}
The NeRF generates a scene corresponding to the provided input image path using learned latents from PC-VAE. This function tests the NeRF decoder's functionality with a given belief state. When the input image is a fully observable view (corresponding to a unknown belief state), the NeRF renders clear actor and ego configurations respecting the input. This allows us to visualize the scene at different timestamps and in different configurations.

\begin{table}
    \begin{tabular*}{0.91\textwidth}{p{0.3\textwidth}P{0.2\textwidth}P{0.2\textwidth}P{0.17\textwidth}}
        \toprule
        Architectures & Train PSNR & SVM Accuracy & NV PSNR \\
        \toprule
        \multicolumn{4}{p{0.91\textwidth}}{\textbf{Multi-Scene Approaching Intersection}} \\
        \midrule
        \rowcolor{gray!10}
        PC-VAE & \textbf{26.47} & \textbf{89.17} & \textbf{26.37}\\
        PC-VAE w/o CL & 26.20 & 83.83 & 26.16 \\
        \rowcolor{gray!10}
        Vanilla PC-VAE & 25.97 & 29.33 & 25.93 \\
        PC-VAE w/o Freezing & 24.82 & 29.83 & 24.78 \\
        \rowcolor{gray!10}% 
        PC-VAE w/ MobileNet & 19.37 & 29.50 & 19.43 \\
        Vanilla VAE & 26.04 & 14.67 & 9.84 \\
        \toprule
        \multicolumn{4}{p{0.91\textwidth}}{\textbf{Multi-Scene Two Lane Merge}} \\
        \midrule
        \rowcolor{gray!10}
        PC-VAE & \textbf{25.50} & \textbf{88.33} & \textbf{25.84}\\
        PC-VAE w/o CL & 24.38 & 29.67 & 24.02 \\
        \rowcolor{gray!10}
        Vanilla PC-VAE & 24.75 & 29.67 & 24.96 \\
        PC-VAE w/o Freezing & 23.97 & 28.33 & 24.04 \\
        \rowcolor{gray!10}% 
        PC-VAE w/ MobileNet & 17.70 & 75.00 & 17.65 \\
        Vanilla VAE & 25.11 & 28.17 & 8.49 \\
        \bottomrule
    \end{tabular*}
    \caption{\textbf{PC-VAE metrics and ablations across Multi-Scene datasets.} CARFF's PC-VAE outperforms other encoder architectures across the Multi-Scene datasets in reconstruction and pose-conditioning.}
    \label{tab:ablations_full}
\end{table}

\section{CARFF Evaluation}
\label{sec:appendix-vae}
\begin{figure}[h]
    \centering
    \begin{subfigure}{0.4\textwidth}
        \centering
        \resizebox{\textwidth}{!}{\begin{tikzpicture}

\definecolor{darkslategray38}{RGB}{38,38,38}
\definecolor{indianred}{RGB}{205,92,92}
\definecolor{lightgray204}{RGB}{204,204,204}
\definecolor{steelblue}{RGB}{70,130,180}
\Large 

\begin{axis}[
axis line style={lightgray204},
tick align=outside,
axis line style={lightgray204},
legend cell align={left},
legend style={
  fill opacity=0.8,
  draw opacity=1,
  text opacity=1,
  at={(0.97,0.03)},
  anchor=south east,
  draw=lightgray204,
},
x grid style={lightgray204},
xlabel=\textcolor{darkslategray38}{Epochs},
xmajorgrids,
xmin=0, xmax=150,
xtick style={color=darkslategray38},
y grid style={lightgray204},
ylabel=\textcolor{darkslategray38}{Average PSNR},
ymajorgrids,
ymin=0, ymax=27,
ytick pos=left,
ytick style={color=darkslategray38}
]

\addplot [line width=2pt, indianred]
table {%
0 3.2337961382865900
1 22.862864776611300
2 23.997979429245000
3 24.344825468063400
4 24.63897071266170
5 24.912699323654200
6 24.963452898025500
7 25.188606634140000
8 25.232166681289700
9 25.442613986969000
10 25.624550830841100
11 25.530961288452100
12 25.568295776367200
13 25.66792795562740
14 25.617633752822900
15 25.662705947876000
16 25.712304954528800
17 25.660220552444500
18 25.671717992782600
19 25.738295944213900
20 25.782612939834600
21 25.861934240341200
22 25.841008172988900
23 25.888316293716400
24 25.869787687301600
25 25.886844867706300
26 25.918680335998500
27 25.82783747291570
28 25.937744358062700
29 26.04253851890560
30 25.94798705291750
31 26.012319522857700
32 25.951027799606300
33 25.945047477722200
34 25.952086082458500
35 26.000554456710800
36 26.038919149398800
37 26.090903457641600
38 26.071060693740800
39 26.09704573059080
40 26.10609062194820
41 26.125283580780000
42 26.134333219528200
43 26.154462339401200
44 26.158104606628400
45 26.131115673065200
46 26.18779100036620
47 26.1387738571167
48 26.205689472198500
49 26.158772859573400
50 26.184632587432900
51 26.15720745277410
52 26.176964311599700
53 26.232639289856000
54 26.189361059188800
55 26.24977097129820
56 26.14964695739750
57 26.268706727981600
58 26.2738058719635
59 26.267332717895500
60 26.274874015808100
61 26.245466287612900
62 26.17753294944760
63 26.2700551071167
64 26.2466882019043
65 26.261005018234300
66 26.294034648895300
67 26.21898489570620
68 26.317392503738400
69 26.296612092971800
70 26.275159685134900
71 26.26576527786260
72 26.260061641693100
73 26.219397449493400
74 26.205199279785200
75 26.222585313797000
76 26.302633632659900
77 26.265934734344500
78 26.319733377456700
79 26.287151809692400
80 26.261070644378700
81 26.299489156723000
82 26.24571813583370
83 26.289304752349900
84 26.289147047042800
85 26.311169778823900
86 26.28203342819210
87 26.307574087142900
88 26.297864320755000
89 26.271872646331800
90 26.35458917236330
91 26.31253109741210
92 26.320315231323200
93 26.322087810516400
94 26.249152889251700
95 26.304606361389200
96 26.343841512680100
97 26.299558012008700
98 26.328413230896000
99 26.244158498764000
100 26.30946629142760
101 26.339351354599
102 26.319589124679600
103 26.304336429595900
104 26.33636326408390
105 26.264677282333400
106 26.301660594940200
107 26.34398536682130
108 26.342318061828600
109 26.239240125656100
110 26.311486248016400
111 26.345051023483300
112 26.245624244689900
113 26.311822355270400
114 26.30821481323240
115 26.235176572799700
116 26.334892446517900
117 26.329406045913700
118 26.331341522216800
119 26.316790744781500
120 26.330744342804000
121 26.325906614303600
122 26.333531934738200
123 26.283708610534700
124 26.28326670265200
125 26.31789917755130
126 26.296589223861700
127 26.33652478981020
128 26.29657049369810
129 26.332081377029400
130 26.281372579574600
131 26.3530165348053
132 26.26218647003170
133 26.324459197998000
134 26.3224371509552
135 26.299309816360500
136 26.2389573097229
137 26.280952388763400
138 26.28218236541750
139 26.263153957366900
140 26.24644997215270
141 26.343030937194800
142 26.33662201309200
143 26.353884365081800
144 26.299360342025800
145 26.346186878204300
146 26.32841129875180
147 26.306835069656400
148 26.31747489929200
149 26.341956699371300
150 26.37695069694520
};

\end{axis}

\end{tikzpicture}}
        \caption{Single-Scene Intersection}
    \end{subfigure}
    \begin{subfigure}{0.4\textwidth}
        \centering
        \resizebox{\textwidth}{!}{\begin{tikzpicture}

\definecolor{darkslategray38}{RGB}{38,38,38}
\definecolor{indianred}{RGB}{205,92,92}
\definecolor{lightgray204}{RGB}{204,204,204}
\definecolor{steelblue}{RGB}{70,130,180}
\Large 

\begin{axis}[
axis line style={lightgray204},
tick align=outside,
axis line style={lightgray204},
legend cell align={left},
legend style={
  fill opacity=0.8,
  draw opacity=1,
  text opacity=1,
  at={(0.97,0.03)},
  anchor=south east,
  draw=lightgray204,
},
x grid style={lightgray204},
xlabel=\textcolor{darkslategray38}{Epochs},
xmajorgrids,
xmin=0, xmax=150,
xtick style={color=darkslategray38},
y grid style={lightgray204},
ylabel=\textcolor{darkslategray38}{Average PSNR},
ymajorgrids,
ymin=0, ymax=27,
ytick pos=left,
ytick style={color=darkslategray38}
]

\addplot [line width=2pt, indianred]
table {%
0 3.67720687687397
1 21.283548113505000
2 22.877644329071000
3 23.61492870012920
4 24.05896824200950
5 24.230868231455500
6 24.534036944707200
7 24.73828126589460
8 24.52697009086610
9 24.96880208015440
10 25.12058069864910
11 25.18389520963030
12 25.18634894371030
13 25.387414779663100
14 25.433933817545600
15 25.506666669845600
16 25.565513671239200
17 25.62898424466450
18 25.724551105499300
19 25.616476491292300
20 25.67451316197710
21 25.773995215098100
22 25.766333560943600
23 25.748972682952900
24 25.7715083916982
25 25.843180150985700
26 25.919112777710000
27 25.878821207682300
28 25.8889479637146
29 25.91290147781370
30 25.95422636985780
31 25.929057200749700
32 25.926606222788500
33 25.375636437734000
34 25.43039834022520
35 25.51186372121180
36 25.58307660738630
37 25.599926935831700
38 25.632146638234500
39 25.79533511161800
40 25.737815030415900
41 25.854072046279900
42 25.91095635096230
43 25.844835414886500
44 25.90539435068770
45 26.0001029364268
46 25.980501823425300
47 26.003353471756000
48 26.059195232391400
49 26.089451910654700
50 26.109266468683900
51 26.059190759658800
52 26.070590012868200
53 26.175123618443800
54 26.129444150924700
55 26.17934103012090
56 26.114924856821700
57 26.18812448501590
58 26.222160142262800
59 26.144255584081000
60 26.182207492192600
61 26.199702262878400
62 26.221144256591800
63 26.213614699045800
64 26.313226598103800
65 26.25955596923830
66 26.295573472976700
67 26.326954288482700
68 26.326193526585900
69 26.278836342493700
70 26.301630109151200
71 26.258643468221000
72 26.366027536392200
73 26.397455450693800
74 26.355214529037500
75 26.313926305770900
76 26.310831947326700
77 26.310831947326700
78 26.35415938059490
79 26.317241605122900
80 26.335534547170000
81 26.39250761350000
82 26.318937419255600
83 26.35498295466110
84 26.40100235303240
85 26.271820160547900
86 26.39549615542090
87 26.394636147817000
88 26.388825168609600
89 26.433420130411800
90 26.389472405115800
91 26.362173833847000
92 26.345164244969700
93 26.322482420603400
94 26.37460387547810
95 26.383501958847000
96 26.416689669291200
97 26.345661808649700
98 26.486547676722200
99 26.3608225663503
100 26.389829037984200
101 26.369706694285100
102 26.371265614827500
103 26.38463985761010
104 26.366175648371400
105 26.357063201268500
106 26.41923297246300
107 26.47067413965860
108 26.381908957163500
109 26.41796510696410
110 26.44475903193160
111 26.412599891026800
112 26.383077920277900
113 26.457307411829600
114 26.391292152404800
115 26.409659277598100
116 26.38514612833660
117 26.43734464963280
118 26.463646694819100
119 26.43378336906430
120 26.43818486849470
121 26.40186970392860
122 26.39402243296310
123 26.41078203201290
124 26.472469008763600
125 26.423410628636700
126 26.37341715812680
127 26.411357475916500
128 26.420434188842800
129 26.493405764897700
130 26.37347099939980
131 26.414342288970900
132 26.40666568438210
133 26.42267261505130
134 26.450110034942600
135 26.51050787607830
136 26.474761902491300
137 26.376424986521400
138 26.409746872584000
139 26.465686696370400
140 26.392600520451900
141 26.422929833730100
142 26.36635331471760
143 26.41899485905970
144 26.46677382469180
145 26.38360802014670
146 26.497042004267400
147 26.38669005393980
148 26.43815764427190
149 26.401059185663900
150 26.46722426732380
};

\end{axis}

\end{tikzpicture}}
        \caption{Multi-Scene Intersection}
    \end{subfigure}
    \begin{subfigure}{0.4\textwidth}
        \centering
        \resizebox{\textwidth}{!}{\begin{tikzpicture}

\definecolor{darkslategray38}{RGB}{38,38,38}
\definecolor{indianred}{RGB}{205,92,92}
\definecolor{lightgray204}{RGB}{204,204,204}
\definecolor{steelblue}{RGB}{70,130,180}
\Large 

\begin{axis}[
axis line style={lightgray204},
tick align=outside,
axis line style={lightgray204},
legend cell align={left},
legend style={
  fill opacity=0.8,
  draw opacity=1,
  text opacity=1,
  at={(0.97,0.03)},
  anchor=south east,
  draw=lightgray204,
},
x grid style={lightgray204},
xlabel=\textcolor{darkslategray38}{Epochs},
xmajorgrids,
xmin=0, xmax=150,
xtick style={color=darkslategray38},
y grid style={lightgray204},
ylabel=\textcolor{darkslategray38}{Average PSNR},
ymajorgrids,
ymin=0, ymax=27,
ytick pos=left,
ytick style={color=darkslategray38}
]

\addplot [line width=2pt, indianred]
table {%
0 3.5041991055011700
1 19.964485913912500
2 21.735898129145300
3 22.73646496772770
4 23.38115350564320
5 23.27671144803370
6 23.81712522347770
7 24.376216247876500
8 24.14866305033370
9 24.305368668238300
10 24.50937872727710
11 24.4797315120697
12 24.440637733141600
13 24.113605343500800
14 24.676989884376500
15 24.649444324175500
16 24.48017028172810
17 24.99414871374770
18 24.602444842656500
19 25.206290402412400
20 24.874020160039300
21 24.980404753685000
22 24.92269426981610
23 25.213535757064800
24 24.96511425177260
25 25.140959405899000
26 25.048650654157000
27 25.17074257214860
28 24.91975162665050
29 25.43163722674050
30 25.356749251683600
31 24.732269988060000
32 25.275322691599500
33 25.227331153551700
34 25.19139658133190
35 25.449872666994700
36 25.479976773262000
37 25.226130695343000
38 25.41928676923120
39 25.298445070584600
40 25.210214049021400
41 25.430085940361000
42 25.530139751434300
43 25.49175564289090
44 25.3972438955307
45 25.46824402332310
46 25.34782070795700
47 25.280229477882400
48 25.38045334815980
49 25.49612751801810
50 25.356165720621700
51 25.515928026835100
52 25.317275018692000
53 25.505737069447800
54 25.394185994466100
55 25.00277036190030
56 25.264367361068700
57 25.162397391001400
58 25.26288625717160
59 25.25251485188800
60 25.251780649821000
61 25.586086254119900
62 25.410022373199500
63 25.444015622139000
64 25.396147367159500
65 25.361280301411900
66 25.208222557703700
67 25.10084145387010
68 25.608635516166700
69 25.511615538597100
70 25.40155162334440
71 25.336510739326500
72 25.65608064810440
73 25.357297239303600
74 25.42470411936440
75 25.351381023724900
76 25.497876586914100
77 25.468961815834000
78 25.159863181114200
79 25.487467606862400
80 25.653418339093500
81 25.42617184003190
82 25.3001827955246
83 25.364122370084100
84 25.383736017545100
85 25.164087295532200
86 25.21738901456200
87 25.502826957702600
88 25.233863986333200
89 25.251440281868000
90 25.689548870722500
91 25.150958847999600
92 25.190545434951800
93 25.47617730776470
94 25.535968267122900
95 25.456829007466600
96 25.256470576922100
97 25.376246040662100
98 25.523263894716900
99 25.58232135295870
100 25.383918231328300
101 25.48218333085380
102 25.119836796124800
103 25.464251244862900
104 25.00395809809370
105 24.984746764500900
106 25.554210292498300
107 25.54588709672290
108 25.08794092496240
109 25.333302777608200
110 25.38168714682260
111 25.393728448549900
112 25.112459594408700
113 25.15163361708320
114 25.796405084927900
115 25.288795024553900
116 25.324369599024500
117 25.324753068288200
118 25.11237644036610
119 25.674293597539300
120 25.16818082968390
121 25.560797940890000
122 25.189537943204200
123 25.47105459690090
124 25.331697233518000
125 25.315383531252500
126 25.356169673601800
127 24.88852658112840
128 25.5957647784551
129 25.48923851966860
130 25.538424549102800
131 25.13831727663680
132 25.56562338034310
133 25.357842756907100
134 25.649914487203000
135 25.467868615786200
136 25.48397708574930
137 25.25542899608610
138 25.72537221590680
139 25.475772558848100
140 25.49480799516040
141 25.18995619614920
142 25.352218518257100
143 25.339763724009200
144 25.270155124664300
145 25.604685815175400
146 25.335854930877700
147 25.657429593404100
148 25.714205872217800
149 25.275484972000100
150 25.24778435866040
};

\end{axis}

\end{tikzpicture}}
        \caption{Multi-Scene Two Lane Merge}
    \end{subfigure}
    \begin{subfigure}{0.4\textwidth}
        \centering
        \resizebox{\textwidth}{!}{\begin{tikzpicture}

\definecolor{darkslategray38}{RGB}{38,38,38}
\definecolor{indianred}{RGB}{205,92,92}
\definecolor{lightgray204}{RGB}{204,204,204}
\definecolor{steelblue}{RGB}{70,130,180}
\Large 

\begin{axis}[
axis line style={lightgray204},
tick align=outside,
axis line style={lightgray204},
legend cell align={left},
legend style={
  fill opacity=0.8,
  draw opacity=1,
  text opacity=1,
  at={(0.97,0.03)},
  anchor=south east,
  draw=lightgray204,
},
x grid style={lightgray204},
xlabel=\textcolor{darkslategray38}{Epochs},
xmajorgrids,
xmin=0, xmax=150,
xtick style={color=darkslategray38},
y grid style={lightgray204},
ylabel=\textcolor{darkslategray38}{Average PSNR},
ymajorgrids,
ymin=0, ymax=30,
ytick pos=left,
ytick style={color=darkslategray38}
]

\addplot [line width=2pt, indianred]
table {%
0 4.004668716071310
1 17.360487378161900
2 19.87633345092550
3 21.6973828094593
4 23.48667598116230
5 24.877374469370100
6 25.885907670725900
7 26.094002654587000
8 26.549829441568100
9 27.01088109223740
10 27.21360444331520
11 27.28381369770440
12 27.37914917434470
13 27.792635599772100
14 27.755080112512600
15 27.706809652024400
16 27.931270405866100
17 27.922996783602100
18 27.770100330960900
19 27.85278978209570
20 28.034636967424000
21 28.273937529411900
22 28.23173918240310
23 28.072072319362500
24 28.194580686265100
25 28.20851685344310
26 28.360984028249500
27 28.24797587463820
28 28.404399415721100
29 28.17111034669740
30 28.305936854818600
31 28.507743890734700
32 28.384433359339600
33 27.8817195892334
34 28.51124083477520
35 28.553848777992100
36 28.216455597808400
37 28.462961058685700
38 28.566133913786500
39 28.335640340611600
40 28.430554417596300
41 28.048356263533900
42 28.384594392085400
43 28.475732402525100
44 28.59683198514190
45 28.47481601825660
46 28.585209915603400
47 28.640362919240800
48 28.530568454576600
49 28.69305880173390
50 28.89296647776730
51 28.58982901642290
52 28.832038644431300
53 28.517594268356500
54 28.500480900640100
55 28.865818493608100
56 28.465305701546000
57 28.776969660883400
58 28.60081801207170
59 28.885517714680100
60 28.790621577829600
61 28.856912391773200
62 28.681534836257700
63 28.829311039136800
64 28.92764217266140
65 28.599189910335800
66 28.268316946167900
67 28.701413126959300
68 28.66631361367050
69 28.82635376418850
70 28.784607831982600
71 28.67555040553000
72 28.993630160456100
73 28.843825298806900
74 28.698636511097800
75 28.76877096424930
76 28.806137416673700
77 28.888911731001300
78 28.879115256710300
79 28.68450221462530
80 28.576155883678500
81 29.102294009664800
82 28.919677582340000
83 29.073174214017600
84 29.040737801703900
85 29.23078321373980
86 28.524685873501500
87 28.639924740445800
88 28.703307967255100
89 28.659975328307200
90 28.909073041833000
91 28.8481155063795
92 28.68402792059860
93 28.732886922532200
94 28.800942103068000
95 28.637162443520400
96 28.74299983701840
97 28.744749400926700
98 28.9032658563144
99 28.681710879008000
100 28.644187623176000
101 28.93745035365010
102 28.86961350924730
103 28.757761747940700
104 28.951228114141900
105 28.784433351046800
106 28.917943180471200
107 28.608015281566700
108 28.749357292617600
109 28.7371690860693
110 28.813858820044500
111 28.74082891491880
112 28.652020592620400
113 29.07293883613920
114 28.876667810523000
115 29.087534130483400
116 28.668387938236800
117 28.75678419030230
118 28.91127466118850
119 28.70741085384200
120 28.704894300820200
121 28.911856222843800
122 28.686075459355900
123 28.78235376053960
124 28.707807223002100
125 28.899795159049700
126 28.837934079377500
127 28.80975029433980
128 28.75554191202360
129 28.83780636994740
130 28.85879421234130
131 28.657860203065700
132 28.743687007738200
133 28.731242995331300
134 28.82303041651630
135 28.58954183606130
136 28.9261067017265
137 28.765645994656300
138 28.561007458230700
139 28.829144947770700
140 29.27582541756010
141 29.106372003969900
142 28.674371705539000
143 28.854661374852300
144 28.600218137105300
145 28.80645507314930
146 28.953605859176000
147 28.69449153844860
148 28.639354774917400
149 28.865503767262300
150 28.804242783698500
};

\end{axis}

\end{tikzpicture}}
        \caption{Hand-manipulation}
    \end{subfigure}
    \caption{\textbf{Average train PSNR plot for all CARLA datasets and the hand-manipulation dataset}. The plot shows the increase in average training PSNR of all images for each dataset, over the period of the training process.}
    \label{fig:psnr_plots}
\end{figure}
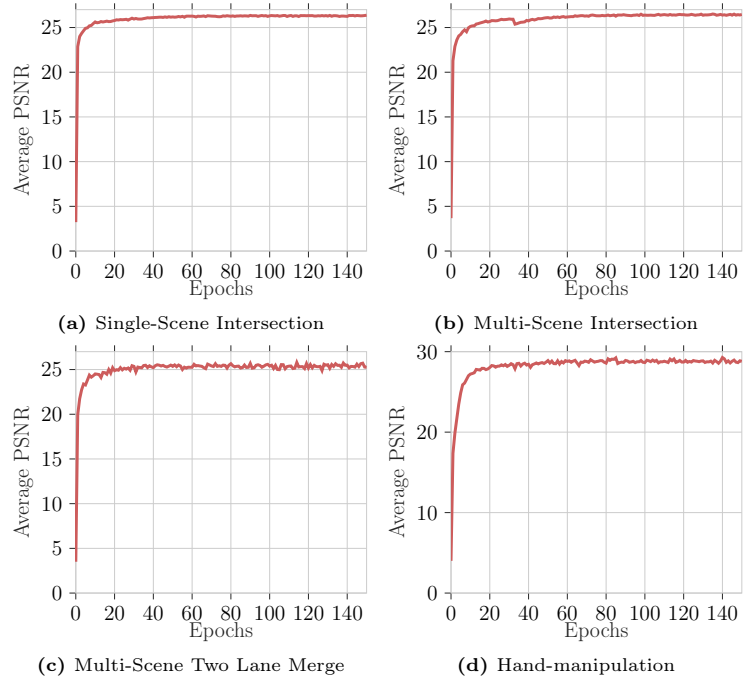
\subsection{Pose-Conditional VAE}

\paragraph{Reconstruction Quality:}
To analyze the reconstruction performance of the model during training, we periodically plot grids of reconstructed images.
These grids consist of (a) randomly selected encoder inputs drawn from the dataset, (b) the corresponding ground truth images for those inputs at each timestamp at the same camera pose, and (c) reconstructed outputs at randomly sampled poses respecting the input scene and timestamp.
An example reconstruction grid is provided in Fig.~\ref{fig:recons}. 
The grid enables visual assessment of whether the model is capable of accurately reconstructing reasonable images using the encoder inputs, conditioned on the poses.
This evaluation provides us with visual evidence of improvement in reconstruction quality. We also quantitatively analyze the progressive improvement of reconstruction through the average PSNR calculated over the training data (see Fig.~\ref{fig:psnr_plots}).

The PC-VAE outputs in Fig.~\ref{fig:recons} only provides visual confirmation to assess the quality of the latents learned by PC-VAE. Utilization of the 3D decoder later in our method allows us to produce more high resolution visualizations of the scene that can be used for further downstream tasks.

\begin{table*}
    \centering
    \begin{tabular}{P{0.11\textwidth}P{0.07\textwidth}P{0.07\textwidth}P{0.07\textwidth}P{0.07\textwidth}P{0.07\textwidth}P{0.07\textwidth}P{0.07\textwidth}P{0.07\textwidth}P{0.07\textwidth}P{0.07\textwidth}}
    \toprule
        \multicolumn{11}{p{0.9\textwidth}}{\textbf{Single-Scene Approaching Intersection}}\\
        Result & $I_{t_1}$ & $I_{t_2}$ & $I_{t_3}$ & $I_{t_4}$ & $I_{t_5}$ & $I_{t_6}$ & $I_{t_7}$ & $I_{t_8}$ & $I_{t_9}$ & $I_{t_{10}}$\\
    \midrule
         $\tilde{I}_{t_1}$ & \textbf{29.01} & \minus 5.97 & \minus 6.08 & \minus 6.52 & \minus 6.44 & \minus 6.03 & \minus 6.31 & \minus 6.36 & \minus 6.26 & \minus 6.28 \\
         $\hat{I}_{t_2}$ & \minus 5.42 & \textbf{27.51} & \minus 3.07 & \minus 4.67 & \minus 4.58 & \minus 4.17 & \minus 4.43 & \minus 4.51 & \minus 4.39 & \minus 4.39 \\
         $\hat{I}_{t_3}$ & \minus 6.06 & \minus 2.81 & \textbf{28.12} & \minus 4.47 & \minus 4.68 & \minus 4.19 & \minus 4.05 & \minus 4.61 & \minus 4.47 & \minus 4.52 \\
         $\hat{I}_{t_4}$ & \minus 7.01 & \minus 5.37 & \minus 5.03 & \textbf{29.40} & \minus 4.99 & \minus 5.08 & \minus 5.03 & \minus 5.41 & \minus 5.28 & \minus 5.32 \\
         $\hat{I}_{t_5}$ & \minus 6.87 & \minus 5.2 & \minus 4.93 & \minus 5.00 & \textbf{29.44} & \minus 4.53 & \minus 4.46 & \minus 5.19 & \minus 5.05 & \minus 5.09 \\
         $\hat{I}_{t_6}$ & \minus 6.29 & \minus 4.55 & \minus 4.27 & \minus 4.8 & \minus 4.24 & \textbf{29.02} & \minus 4.02 & \minus 4.53 & \minus 4.38 & \minus 4.44 \\
         $\hat{I}_{t_7}$ & \minus 6.76 & \minus 5.05 & \minus 4.76 & \minus 5.31 & \minus 5.14 & \minus 4.36 & \textbf{29.50} & \minus 4.50 & \minus 4.86 & \minus 4.93 \\
         $\hat{I}_{t_8}$ & \minus 6.73 & \minus 5.02 & \minus 4.74 & \minus 5.25 & \minus 5.10 & \minus 4.64 & \minus 4.76 & \textbf{29.46} & \minus 4.41 & \minus 4.86 \\
         $\hat{I}_{t_9}$ & \minus 6.75 & \minus 5.00 & \minus 4.70 & \minus 5.23 & \minus 5.07 & \minus 4.64 & \minus 4.85 & \minus 4.52 & \textbf{29.55} & \minus 4.42 \\
         $\hat{I}_{t_{10}}$ & \minus 6.79 & \minus 5.06 & \minus 4.75 & \minus 5.30 & \minus 5.15 & \minus 4.69 & \minus 4.93 & \minus 5.01 & \minus 4.34 & \textbf{29.55} \\
    \end{tabular}
    \begin{tabular}{P{0.11\textwidth}P{0.12\textwidth}P{0.12\textwidth}P{0.12\textwidth}P{0.12\textwidth}P{0.12\textwidth}P{0.12\textwidth}}
    \toprule
        \multicolumn{7}{p{0.9\textwidth}}{\textbf{Multi-Scene Approaching Intersection}} \\
        Result & $I_{t_1}$ & $I_{t_2}$ & $I_{t_3}$ & $I_{t_4}$ & $I_{t_5}$ & $I_{t_6}$ \\
    \midrule
         $\tilde{I}_{t_1}$ & \textbf{28.10} & \minus 5.24 & \minus 5.50 & \minus 1.67 & \minus 3.29 & \minus 3.92 \\
         $\hat{I}_{t_2}$ & \minus 5.23 & \textbf{28.02} & \minus 6.11 & \minus 4.70 & \minus 3.21 & \minus 4.84 \\
         $\hat{I}_{t_3}$ & \minus 5.43 & \minus 6.03 & \textbf{27.97} & \minus 4.85 & \minus 4.53 & \minus 2.93 \\
         $\tilde{I}_{t_4}$ & \minus 1.71 & \minus 4.73 & \minus 5.00 & \textbf{28.26} & \minus 2.25 & \minus 3.08 \\
         $\tilde{I}_{t_5}$ & \minus 3.68 & \minus 3.24 & \minus 4.91 & \minus 2.76 & \textbf{28.21} & \minus 2.99 \\
         $\hat{I}_{t_6}$ & \minus 4.02 & \minus 4.91 & \minus 3.27 & \minus 3.13 & \minus 2.61 & \textbf{28.26} \\
    \toprule
        \multicolumn{7}{p{0.9\textwidth}}{\textbf{Multi-Scene Two Lane Merge}} \\
        Result & $I_{t_1}$ & $I_{t_2}$ & $I_{t_3}$ & $I_{t_4}$ & $I_{t_5}$ & $I_{t_6}$ \\
    \midrule
         $\tilde{I}_{t_1}$ & \textbf{28.27} & \minus 5.31 & \minus 6.41 & \textbf{28.23} & \minus 4.77 & \minus 5.42 \\
         $\tilde{I}_{t_2}$ & \minus 5.22 & \textbf{28.23} & \minus 5.17 & \minus 5.27 & \minus 2.91 & \minus 4.01 \\
         $\hat{I}_{t_3}$ & \minus 6.32 & \minus 5.09 & \textbf{28.14} & \minus 6.33 & \minus 5.01 & \minus 4.28 \\
         $\tilde{I}_{t_4}$ & \textbf{28.27} & \minus 5.27 & \minus 6.37 & \textbf{28.23} & \minus 4.72 & \minus 5.37 \\
         $\tilde{I}_{t_5}$ & \minus 4.64 & \minus 2.73 & \minus 5.01 & \minus 4.71 & \textbf{28.08} & \minus 5.29 \\
         $\hat{I}_{t_6}$ & \minus 5.32 & \minus 4.02 & \minus 4.32 & \minus 5.33 & \minus 5.34 & \textbf{28.17} \\
    \toprule
        \multicolumn{7}{p{0.9\textwidth}}{\textbf{Hand-manipulation Object Reaching}} \\
        Result & $I_{t_1}$ & $I_{t_2}$ & $I_{t_3}$ & $I_{t_4}$ & $I_{t_5}$ & $I_{t_6}$ \\
    \midrule
         $\tilde{I}_{t_1}$ & \textbf{30.71} & \minus 10.74 & \minus 10.63 & \textbf{30.71} & \minus 9.81 & \minus 11.13 \\
         $\tilde{I}_{t_2}$ & \minus 10.07 & \textbf{30.32} & \minus 9.25 & \minus 10.07 & \minus 10.88 & \minus 10.17 \\
         $\hat{I}_{t_3}$ & \minus 9.98 & \minus 9.11 & \textbf{30.37} & \minus 9.98 & \minus 10.78 & \minus 10.02 \\
         $\tilde{I}_{t_4}$ & \textbf{30.66} & \minus 10.69 & \minus 10.58 & \textbf{30.66} & \minus 9.77 & \minus 11.09 \\
         $\tilde{I}_{t_5}$ & \minus 8.19 & \minus 10.21  & \minus 10.1 & \minus 8.19 & \textbf{29.49} & \minus 9.99 \\
         $\hat{I}_{t_6}$ & \minus 10.96 & \minus 10.6 & \minus 10.49 & \minus 10.96 & \minus 11.08 & \textbf{30.67} \\
    \bottomrule
    \end{tabular}
    \caption{\textbf{Complete PSNR values for fully observable predictions for all CARLA datasets and the Hand-manipulation dataset.} The table contains PSNR values between the ground truth images and either a toggled image (marked as $\tilde{I}_{t_i}$), or a predicted image from the NeRF decoder (marked as $\hat{I}_{t_i}$). Toggled or predicted images that correspond to the correct ground truth are bolded and have a extremely high PSNR value, indicating high fidelity results. The PSNR values for incorrect correspondances are replaced with the difference between the incorrect PSNR and the bolded PSNR associated with a correct correspondance.}
    \label{tab:predictions_whole}
\end{table*}

\paragraph{Latent Space Analysis}
To assess the quality of the latents generated by PC-VAE, we initially use t-SNE plots to visualize the latent distributions as clusters. 
Fig.~\ref{fig:clustering} shows that the distributions of the latent samples for the Multi-Scene Two Lane Merge dataset are separable. 
While t-SNE is good at retaining nearest-neighbor information by preserving local structures, it performs weakly in preserving global structures.
Therefore, t-SNE may be insufficient in capturing the differences in distributions for all our datasets.

Instead, we pivot to Support Vector Machine to perform a quantitative evaluation of the separability of the latents. We utilize a Radial Basis Function (RBF) kernel with the standard regularization parameter ($C = 1$). 
We perform 10-fold validation on the latents to calculate the accuracy as a metric for clustering. See Tab.~\ref{tab:ablations_full} for the results. 

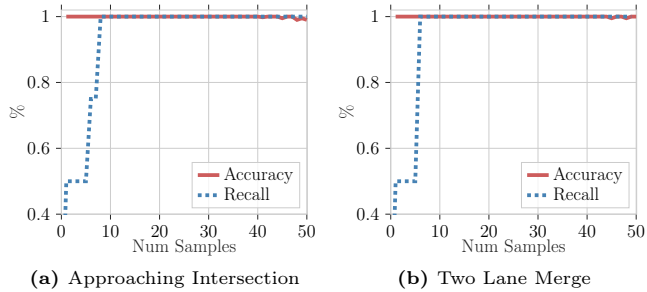
\begin{figure}
    \centering
    \begin{subfigure}{.35\textwidth} % Adjust the width as needed
        \centering
        \resizebox{\textwidth}{!}{\begin{tikzpicture}

\definecolor{darkslategray38}{RGB}{38,38,38}
\definecolor{indianred}{RGB}{205,92,92}
\definecolor{lightgray204}{RGB}{204,204,204}
\definecolor{steelblue}{RGB}{70,130,180}
\Large 

\begin{axis}[
axis line style={lightgray204},
tick align=outside,
axis line style={lightgray204},
legend cell align={left},
legend style={
  fill opacity=0.8,
  draw opacity=1,
  text opacity=1,
  at={(0.97,0.03)},
  anchor=south east,
  draw=lightgray204,
},
x grid style={lightgray204},
xlabel=\textcolor{darkslategray38}{Num Samples},
xmajorgrids,
xmin=0, xmax=50.1,
xtick style={color=darkslategray38},
y grid style={lightgray204},
ylabel=\textcolor{darkslategray38}{\%},
ymajorgrids,
ymin=0.4, ymax=1.02,
ytick pos=left,
ytick style={color=darkslategray38}
]

\addlegendentry{Accuracy}
\addplot [line width=3pt, indianred]
table {%
1 1.0
2 1.0
3 1.0
4 1.0
5 1.0
6 1.0
7 1.0
8 1.0
9 1.0
10 1.0
11 1.0
12 1.0
13 1.0
14 1.0
15 1.0
16 1.0
17 1.0
18 1.0
19 1.0
20 1.0
21 1.0
22 1.0
23 1.0
24 1.0
25 1.0
26 1.0
27 1.0
28 1.0
29 1.0
30 1.0
31 1.0
32 1.0
33 1.0
34 1.0
35 1.0
36 1.0
37 1.0
38 1.0
39 1.0
40 1.0
41 0.998
42 1.0
43 1.0
44 1.0
45 0.995
46 1.0
47 1.0
48 0.99
49 0.995
50 0.99
};
\addlegendentry{Recall}

\addplot [line width=3pt, dashed, steelblue]
table {%
0 0
1 0.5
2 0.5
3 0.5
4 0.5
5 0.5
6 0.75
7 0.75
8 1.0
9 1.0
10 1.0
11 1.0
12 1.0
13 1.0
14 1.0
15 1.0
16 1.0
17 1.0
18 1.0
19 1.0
20 1.0
21 1.0
22 1.0
23 1.0
24 1.0
25 1.0
26 1.0
27 1.0
28 1.0
29 1.0
30 1.0
31 1.0
32 1.0
33 1.0
34 1.0
35 1.0
36 1.0
37 1.0
38 1.0
39 1.0
40 1.0
41 1.0
42 1.0
43 1.0
44 1.0
45 1.0
46 1.0
47 1.0
48 1.0
49 1.0
50 1.0
};

\end{axis}

\end{tikzpicture}}
        \caption{Approaching Intersection}
    \end{subfigure}
    \begin{subfigure}{.35\textwidth} % Adjust the width as needed
        \centering
        \resizebox{\textwidth}{!}{\begin{tikzpicture}

\definecolor{darkslategray38}{RGB}{38,38,38}
\definecolor{indianred}{RGB}{205,92,92}
\definecolor{lightgray204}{RGB}{204,204,204}
\definecolor{steelblue}{RGB}{70,130,180}
\Large 

\begin{axis}[
axis line style={lightgray204},
tick align=outside,
axis line style={lightgray204},
legend cell align={left},
legend style={
  fill opacity=0.8,
  draw opacity=1,
  text opacity=1,
  at={(0.97,0.03)},
  anchor=south east,
  draw=lightgray204,
},
x grid style={lightgray204},
xlabel=\textcolor{darkslategray38}{Num Samples},
xmajorgrids,
xmin=0, xmax=50.1,
xtick style={color=darkslategray38},
y grid style={lightgray204},
ylabel=\textcolor{darkslategray38}{\%},
ymajorgrids,
ymin=0.4, ymax=1.02,
ytick pos=left,
ytick style={color=darkslategray38}
]

\addlegendentry{Accuracy}
\addplot [line width=3pt, indianred]
table {%
1 1.0
2 1.0
3 1.0
4 1.0
5 1.0
6 1.0
7 1.0
8 1.0
9 1.0
10 1.0
11 1.0
12 1.0
13 1.0
14 1.0
15 1.0
16 1.0
17 1.0
18 1.0
19 1.0
20 1.0
21 1.0
22 1.0
23 1.0
24 1.0
25 1.0
26 1.0
27 1.0
28 1.0
29 1.0
30 1.0
31 1.0
32 1.0
33 1.0
34 1.0
35 1.0
36 1.0
37 1.0
38 1.0
39 1.0
40 1.0
41 1.0
42 1.0
43 1.0
44 1.0
45 0.995
46 1.0
47 1.0
48 0.995
49 1.0
50 1.0
};

\addlegendentry{Recall}
\addplot [line width=3pt, dashed, steelblue]
table {%
0 0
1 0.5
2 0.5
3 0.5
4 0.5
5 0.5
6 1
7 1.0
8 1.0
9 1.0
10 1.0
11 1.0
12 1.0
13 1.0
14 1.0
15 1.0
16 1.0
17 1.0
18 1.0
19 1.0
20 1.0
21 1.0
22 1.0
23 1.0
24 1.0
25 1.0
26 1.0
27 1.0
28 1.0
29 1.0
30 1.0
31 1.0
32 1.0
33 1.0
34 1.0
35 1.0
36 1.0
37 1.0
38 1.0
39 1.0
40 1.0
41 1.0
42 1.0
43 1.0
44 1.0
45 1.0
46 1.0
47 1.0
48 1.0
49 1.0
};

\end{axis}

\end{tikzpicture}}
        \caption{Two Lane Merge}
    \end{subfigure}
    \caption{\textbf{Multi-Scene dataset accuracy and recall curves from learned latents.} We test our framework across $n = 1$ and $n = 50$ samples from PC-VAE's latent distributions from ego-centric image input. Across the number of samples $n$, we achieve an ideal margin of belief state coverage generated under partial observation (recall), and the proportion of correct beliefs sampled under full observation (accuracy) for the MDN to learn. As we significantly increase the number of samples, the accuracy starts to decrease due to randomness in latent distribution resampling.}
    \label{fig:learned_curves}
\end{figure}

Beyond separability, we analyze the recall and accuracy of the learned latents directly from PC-VAE under partial and full observations. This achieves very high accuracy even under a large number of samples while retraining decent recall, enabling downstream MDN training. (See Fig.~\ref{fig:learned_curves})

For the additional Hand-manipulation Object Reaching dataset, we used a similar setup as detailed in Fig~\ref{fig:curves} with $n = 1$ to $n = 50$ samples from the MDN's predicted latent distributions from an potentially partially observable image input. Similar to the CARLA dataset results, we achieve an ideal margin of belief state coverage generated under partial observation (recall), and the proportion of correct beliefs sampled under full observation (accuracy).

\subsection{Fully Observable Predictions}
One of the tasks of the MDN is to forecast the future scene configurations under full observation. 
We quantitatively evaluate our model's ability to forecast future scenes by comparing bird's-eye views rendered from the NeRF with chosen ground truth images of the scene for the various timestamps (see Tab.~\ref{tab:predictions_whole}). The values are calculated and displayed for all four datasets. In Tab.~\ref{tab:predictions_whole}, images are marked as either toggled ($\tilde{I}_{t_i}$) or predicted ($\hat{I}_{t_i}$). Toggled images in the table cannot be predicted deterministically due to it being the first timestamp in the dataset, or the state of the previous timestamps across scenes being the same in case of dynamics uncertainty. Due to the same reason, in the Multi-Scene Two Lane Merge and the Hand-manipulation Object Reaching Datasets, there are additional bolded PSNR values for the pairs $(I_{t_1}, \tilde{I}_{t_4})$ and $(I_{t_4}, \tilde{I}_{t_1})$.

We demonstrate that the toggled or predicted images that correspond to the correct ground truth show a PSNR value around $29$, indicating high fidelity 3D reconstruction and clear visual decodings as the output of \oursshort{}.

\subsection{Architecture Ablations}
\label{sec:appendix-ablations}
The results presented in Tab.~\ref{tab:ablations} substantiate the benefits of our PC-VAE encoder architecture compared to other formulations. Specifically, a non-conditional VAE fails in SVM accuracy as it only reconstructs images and does not capture the underlying 3D structures. Vanilla PC-VAE and PC-VAE without freezing weights require careful fine-tuning of several hyper-parameters and don't generalize well to drastic camera movements. Our experiments show that our proposed model is capable of sustaining stochastic characteristics via latent representations in the presence of occlusion, while simultaneously ensuring precise reconstructions.

\begin{table*}
\begin{tabular*}{0.85\textwidth}{m{0.28\textwidth}P{0.18\textwidth}P{0.18\textwidth}P{0.18\textwidth}} 
        \toprule
        Encoder \hspace{2mm} Architectures & Train PSNR & SVM Acc. & NV PSNR \\ \midrule
        \rowcolor{gray!10}
        PC-VAE & \textbf{26.30} & \textbf{75.20} & \textbf{25.24}\\
        PC-VAE w/o CL & 26.24 & 70.60 & 24.80\\
        \rowcolor{gray!10}
        Vanilla PC-VAE & 26.02 & 25.70 & 24.65 \\
        PC-VAE w/o Freezing & 24.57 & 5.80 & 24.60\\
        \rowcolor{gray!10}% 
        PC-VAE w/ MobileNet & 17.14 & 19.70 & 17.16 \\
        \midrule
        Vanilla VAE & 24.15 & 10.60 & 11.43 \\
        \bottomrule
    \end{tabular*}
\caption{\textbf{PC-VAE ablations}. CARFF's PC-VAE encoder outperforms other architectures in image reconstruction and pose-conditioning. We evaluate on: PC-VAE without Conv. Layer, PC-VAE with a vanilla encoder, PC-VAE without freezing ViT weights, PC-VAE replacing ViT with MobileNet, and non pose-conditional VAE. }
  \label{tab:ablations}
\end{table*}

\bibliographystyle{splncs04}
\bibliography{main}
\end{document}